\PassOptionsToPackage{table}{xcolor}
\documentclass[sigconf, screen]{acmart}
\usepackage{graphicx}
\usepackage{booktabs}
\usepackage{diagbox}
\usepackage{makecell}
\usepackage{pifont}
\usepackage{xcolor}
\usepackage[most]{tcolorbox}
\usepackage{tabularx}
\usepackage{array}
\usepackage{adjustbox}
\usepackage{newfloat}
\usepackage{listings}
\usepackage{xspace}
\usepackage{mathrsfs}
\usepackage{microtype}

\definecolor{headBlue}{HTML}{D0E3FA}
\definecolor{appleGray}{HTML}{FFF3E0}
\definecolor{appleBlue}{HTML}{F3E5F5}
\definecolor{appleGreen}{HTML}{E9F7EF}

\hypersetup{
  colorlinks=true,
  allcolors=blue
}

\title{\texttt{WorldGenBench}: A World-Knowledge-Integrated Benchmark for Implicit Reasoning-Driven Text-to-Image Generation}

\author{Ruoshi Xu}
\authornote{These authors contributed equally to this work.}
\affiliation{\institution{Southern University of Science and Technology}\country{China}}
\email{xurs2022@email.sustech.edu.cn}

\author{Daoan Zhang}
\authornotemark[1]
\affiliation{\institution{University of Rochester}\country{USA}}
\email{daoan.zhang@rochester.edu}

\author{Che Jiang}
\authornotemark[1]
\affiliation{\institution{Southern University of Science and Technology}\country{China}}
\email{12210914@email.sustech.edu.cn}

\author{Biaoxiang Chen}
\authornotemark[1]
\affiliation{\institution{Southern University of Science and Technology}\country{China}}
\email{12112202@email.sustech.edu.cn}

\author{Zijian Jin}
\affiliation{\institution{New York University}\country{USA}}
\email{zj2076@nyu.edu}

\author{Yutian Lu}
\affiliation{\institution{Datawhale}\country{China}}
\email{physicoad@gmail.com}

\author{Jianguo Zhang}
\affiliation{\institution{Southern University of Science and Technology}\country{China}}
\email{zhangj@sustech.edu.cn}

\author{Liang Yong}
\affiliation{\institution{Chinese Medicine Guangdong Laboratory}\country{China}}
\email{yongliangresearch@gmail.com}

\author{Jiebo Luo}
\authornote{Corresponding authors.}
\affiliation{\institution{University of Rochester}\country{USA}}
\email{jluo@cs.rochester.edu}

\author{Shengda Luo}
\authornotemark[2]
\affiliation{\institution{Chinese Medicine Guangdong Laboratory}\country{China}}
\email{luosd@sustech.edu.cn}



\begin{document}
\begin{abstract}
Recent text-to-image (T2I) models achieve high visual fidelity and accurate explicit grounding from input prompts. 
However, image generation is fundamentally a process of semantic expansion. Beyond surface-level descriptions, generating authentic visual content requires supplementing prompts with unstated cues, a crucial step that current models frequently overlook. 
In response, we introduce WorldGenBench, a benchmark designed to evaluate how well T2I systems integrate factual grounding with contextual inference.
Unlike prior work that isolates these abilities, WorldGenBench emphasizes their synergy, recognizing that true understanding requires both knowing and reasoning. 
To enable interpretable evaluation, we propose the Knowledge Checklist Score, which assesses whether generated images faithfully reflect key explicit and implicit semantic elements. 
Evaluating 24 models reveals persistent gaps: proprietary models like GPT-4o show strong integration, while most open-source systems lag, particularly in culturally diverse and scientific scenarios. Our analysis further identifies common failure modes, including semantic hallucination and the underrepresentation of non-Western knowledge, underscoring the limitations of current models in achieving real-world readiness. These findings highlight the need to guide future T2I models toward deeper, knowledge-aware generation. Project Page: \href{https://dwanzhang-ai.github.io/WorldGenBench/}{https://dwanzhang-ai.github.io/WorldGenBench/}
\end{abstract}

\maketitle

\section{Introduction}
Modern text-to-image (T2I) models exhibit exceptional proficiency in explicit grounding~\cite{xu2023imagereward, ding2021cogview, flux2024, zhang2024seppo}. 
They successfully parse surface-level descriptions and retrieve the corresponding world knowledge to render highly realistic visuals. 
Nevertheless, substantial gaps persist in advancing T2I generation toward genuine \textbf{real-world readiness}. 
Because natural language prompts are inherently under-specified, generating images is fundamentally a process of semantic expansion. 
To synthesize real-world-aligned visual content, models must move beyond literal text-to-image mapping and actively supplement the prompt with unstated visual cues. 
Crucially, rather than being arbitrarily generated, these unstated cues must be logically deduced from the explicitly retrieved facts.
This highlights a critical limitation of current systems, revealing that while explicitly triggered \textbf{world knowledge} provides a necessary factual foundation, it is insufficient on its own.
To bridge this gap, T2I systems must seamlessly integrate broad world knowledge with \textbf{implicit reasoning}, the fundamental cognitive mechanism that derives these unstated but contextually necessary details based on the established facts.

\label{sec:intro}

\begin{figure}[ht]
    \centering
    \includegraphics[width=0.95\linewidth]{images/main260402-v2.pdf}
    \caption{The presence of world knowledge and the emergence of implicit reasoning capabilities are fundamental to building a high-quality text-to-image model, and are essential prerequisites for achieving \textit{real-world ready} generation.}
    \label{fig:1}
\end{figure}

\textbf{World knowledge} serves as the indispensable factual basis, spanning cultural, historical, and scientific domains, which is required to faithfully instantiate concepts ranging from Victorian architecture to polar landscapes. 
Beyond mere retrieval, \textbf{implicit reasoning} must seamlessly apply and extend these established facts within a specific context. While current systems can readily infer basic physical commonsense, such as depicting wet ground and overcast skies for a rainy soccer match, they falter when prompts demand a deeper synergy of facts and logic. As illustrated in Figure~\ref{fig:1}, rendering an authentic image of a farmer in the Bamiyan Valley in the "late fall of 2001" requires the model to not only utilize the cultural knowledge of traditional Hazara clothing but also logically deduce that the background must feature the destroyed Buddha niches rather than the intact statues. 
This intricate interdependence demonstrates why explicit pattern matching fails, and why true real-world readiness hinges on the fusion of knowledge and reasoning.

The integration of \textbf{world knowledge} and \textbf{implicit reasoning} represents an interdependent process rather than a collection of separable functions. Knowledge alone remains insufficient without the cognitive mechanisms required to properly contextualize it within a scene. As the Bamiyan Valley example demonstrates, historical facts are only effective if the model can logically align them with the temporal constraints implied by the prompt. This fundamental synergy is what underpins a T2I system’s capacity to truly comprehend and execute complex user intent in real-world scenarios.

Building upon these insights, we identify the integration of expansive world knowledge and profound implicit reasoning as the central challenge in achieving truly robust and intelligent T2I generation. These dual pillars constitute the fundamental requirement for a model to transcend surface-level pattern matching and genuinely comprehend complex user intent. Recognizing this synergy as the primary technical bottleneck implies that current evaluation paradigms, which often treat knowledge retrieval and logical inference as isolated functions, fail to capture the full complexity required for real-world scene synthesis.

Recent efforts have highlighted the need to evaluate world knowledge and inferential reasoning in T2I generation. Typically, these benchmarks operate by constructing prompts that necessitate world knowledge grounding and inferential reasoning, followed by employing a VLM-as-a-judge to score the generated outputs. For instance, WISE~\cite{niu2025wise} centers on explicit grounding, where the relevant fact is already stated or directly embedded in the prompt, such as when the prompt “Einstein’s favorite musical instrument” simply requires answering “violin”, rather than requiring the model to infer unstated information. PhyBench~\cite{meng2024phybench} concentrates narrowly on physical commonsense, neglecting broader domains like history, culture, and geography. More recently, MMMG~\cite{luo2025mmmg} and R2I-Bench~\cite{chen2025r2i} expanded benchmark scale and dimensionality, but prioritize evaluation quantity over diagnostic depth. MMMG, despite its large size and use of knowledge graphs, suffers from construction bias arising from its template and KG-driven question generation, and lacks implicit reasoning assessment. R2I-Bench includes reasoning prompts but lacks culturally diverse world knowledge coverage.

Motivated by these gaps, we introduce \textbf{WorldGenBench}, the first benchmark to jointly assess world knowledge grounding and implicit reasoning. 
WorldGenBench integrates explicit grounding and implicit reasoning through a diagnostic, visually verifiable checklist that covers both anchored facts and unstated inferences.
Unlike prior work that targets narrow domains~\cite{meng2024phybench, fu2024commonsense} or simply scales prompt volume~\cite{luo2025mmmg}, WorldGenBench spans a broad spectrum of knowledge across the humanities and sciences. Crucially, our human explicitly target general world knowledge (including information about all countries globally) and the ability to infer unstated but accurate information (i.e., whether the model can generate facts not mentioned in the prompt but that are nevertheless accurate).
As summarized in Table~\ref{tab:0}, WorldGenBench is the only benchmark that jointly satisfies all four criteria: world knowledge, implicit reasoning, knowledge diversity, and checklist-based evaluation.

Prior benchmarks typically rely on subjective scoring from VLMs. However, VLM-based evaluation suffers from overgeneralization, vulnerability to score hacking, and insensitivity to factual errors or missing inferences. These issues stem from VLMs' tendency to prioritize surface-level alignment and associative matching over semantic rigor, making it hard to accurately diagnose deficiencies in world knowledge grounding and implicit reasoning.

To address this, we introduce the \textbf{Knowledge Checklist Score}, a structured evaluation method that associates each prompt with a curated checklist of factual and inferential elements. Evaluation measures how many checklist items are accurately realized in the image, enforcing explicit visual verification. This approach mitigates hallucinations, associative bias, and shallow correlations, offering a more objective and interpretable assessment of T2I models' capabilities.

We benchmark 24 state-of-the-art T2I models on WorldGenBench, including 21 open-source and 3 proprietary models. Among the open-source group, we distinguish 8 diffusion-based and 13 auto-regressive models. Results show proprietary models, particularly GPT-4o, outperform others significantly. Diffusion-based models also consistently surpass current auto-regressive ones, highlighting the benchmark’s ability to reveal both broad and subtle model limitations.

Our contributions are summarized as follows:

\begin{itemize}
    \item We introduce \textbf{WorldGenBench}, a new benchmark for evaluating world knowledge grounding and implicit reasoning in text-to-image generation, with broad domain and cultural coverage.
    \item We propose the \textbf{Knowledge Checklist Score}, a structured, fine-grained evaluation metric that mitigates VLM biases and enables reliable diagnosis of model capabilities.
    \item We benchmark 24 state-of-the-art models and reveal significant gaps in current T2I systems, underscoring the need for a deeper understanding beyond surface-level text-image alignment.
\end{itemize}

\begin{table*}
    \centering
    \begin{tabular}{c|c|c|c|c}
        \toprule
        Benchmark & \makecell{World\\Knowledge} & \makecell{Implicit\\Reasoning} & \makecell{Knowledge\\Diversity} & \makecell{Checklist\\Score}  \\
        \hline
        Imagine-E\cite{lei2025imagine} & \textcolor{red}{\ding{55}} & \textcolor{red}{\ding{55}} & \textcolor{red}{\ding{55}} & \textcolor{red}{\ding{55}}  \\
        
        \hline
        PhyBench\cite{meng2024phybench} & \textcolor{red}{\ding{55}} & \textcolor{green}{\ding{51}} &\textcolor{red}{\ding{55}} & \textcolor{red}{\ding{55}}  \\
        \hline
        Commonsense-T2I\cite{fu2024commonsense} & \textcolor{green}{\ding{51}} & \textcolor{red}{\ding{55}} & \textcolor{red}{\ding{55}} & \textcolor{red}{\ding{55}}  \\
        \hline
        
        WISE\cite{niu2025wise} & \textcolor{green}{\ding{51}} & \textcolor{red}{\ding{55}}&\textcolor{green}{\ding{51}} & \textcolor{red}{\ding{55}}  \\
        \hline
        MMMG\cite{luo2025mmmg}& \textcolor{green}{\ding{51}} & \textcolor{red}{\ding{55}}&\textcolor{red}{\ding{55}} & \textcolor{red}{\ding{55}}  \\
        \hline
        R2I-Bench\cite{chen2025r2i} & \textcolor{green}{\ding{51}} & \textcolor{red}{\ding{55}}&\textcolor{red}{\ding{55}} & \textcolor{red}{\ding{55}}  \\
        \hline
        WorldGenBench & \textcolor{green}{\ding{51}} & \textcolor{green}{\ding{51}} &\textcolor{green}{\ding{51}}& \textcolor{green}{\ding{51}} \\ 
        
        \bottomrule
    \end{tabular}
    \caption{Comparison of the proposed WorldGenBench with existing T2I benchmarks.  }
    \label{tab:0}
    \vspace{-6mm}
\end{table*}

\section{Related Work}
\subsection{Evaluation of Text-to-Image Models}

Traditional evaluation of T2I models has primarily focused on image realism and text-image alignment, relying on metrics such as FID \cite{heusel2017gans} and CLIPScore \cite{hessel2021clipscore}. While these metrics fall short in assessing a model’s capacity to understand and apply real-world knowledge. Recent benchmarks such as Commonsense-T2I~\cite{fu2024commonsense}, PhyBench~\cite{meng2024phybench}, WISE~\cite{niu2025wise}, {MMMG}~\cite{luo2025mmmg}, {R2I-Bench}~\cite{chen2025r2i}, LongT2IBench~\cite{yang2025longt2ibench}, SpatialGenEval~\cite{wang2026spatialgeneval} have introduced more complex evaluation scenarios involving compositionality, commonsense, physical reasoning, and spatial awareness. Additionally, TIFA~\cite{hu2023tifa} and T2I-FactualBench~\cite{huang2025factual} evaluate factual consistency and reproduction of knowledge entity. However, these benchmarks focus on \textbf{explicit} reasoning or simply increasing the variety and amount of cases, without delving into the fundamental issues of T2I models, thereby limiting their ability to reflect the nuanced and often \textbf{implicit} knowledge demands present in real-world applications. In contrast, WorldGenBench explicitly proposes that the integration of world knowledge and implicit reasoning is central to T2I, and constructs the benchmark based on this core principle.

\subsection{Text-to-image Models}

T2I generation has progressed along two main lines: dedicated generative models and unified multimodal systems. Early T2I models adopt autoregressive strategies, treating image generation as sequential token prediction, but suffer from high computational costs and limited fidelity. Subsequently, diffusion-based models, such as GLIDE~\cite{nichol2022glidephotorealisticimagegeneration} and early Latent Diffusion Models (LDMs)~\cite{rombach2022highresolutionimagesynthesislatent}, came to dominate the field by generating high-quality images via iterative denoising, predominantly relying on U-Net architectures.

To further improve scalability and semantic alignment, recent advancements have marked a paradigm shift from CNN-based backbones to Diffusion Transformer (DiT) architectures. Building upon this trajectory, the latest Stable Diffusion models~\cite{esser2024scalingrectifiedflowtransformers} and Flux series~\cite{labs2024flux, labs2025flux1kontext} leverage transformer architectures to significantly enhance controllability and generation quality. Following this trend, Qwen-Image~\cite{wu2025qwenimage} leverages multimodal diffusion transformers to achieve deep alignment between text and visual latents, while Z-Image~\cite{imageteam2025zimage} emphasizes extreme step distillation for high-fidelity generation. 

Beyond dedicated T2I models, recent unified multimodal models integrate vision and language within a single framework, enabling both cross-modal understanding and generation. These models~\cite{wang2025simplear,fang2025got,lumina-mgpt-2.0,wu2025harmon,xie2024showo,wang2024Emu3,wu2024janus,chen2025janus,ma2024janusflow,zhang2026cheers,yang2026fuxiuni} extend large language models with visual capabilities via discrete token decoders~\cite{oord2018neuraldiscreterepresentationlearning} or diffusion modules. Notably, Show-O~\cite{xie2024showo} and Kosmos-1~\cite{huang2023languageneedaligningperception} unify bidirectional generation and reasoning across modalities without modality-specific tuning. We evaluate these representative approaches in our study.
\section{WorldGenBench}

\begin{figure}[htbp]
    \centering
    \vspace{-6mm}
    \includegraphics[width=0.8\linewidth]{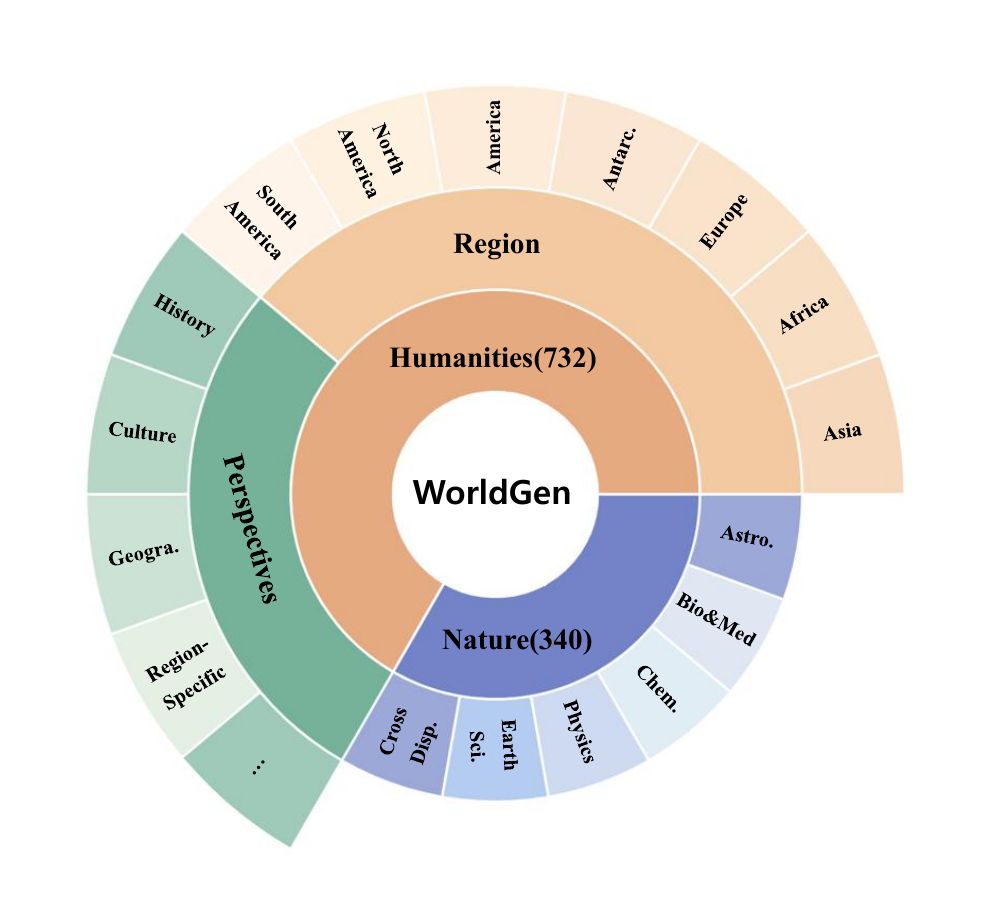}
    \caption{WorldGenBench: Humanities and Nature prompts by region and category.}
    \label{fig:stat}
    \vspace{-5mm}
\end{figure}

WorldGenBench is a benchmark designed to systematically evaluate T2I generation models from the perspectives of world knowledge grounding and implicit reasoning. As shown in Figure.~\ref{fig:stat}, it consists of 1072 carefully curated prompts, each paired with a fine-grained Knowledge Checklist. WorldGenBench covers two major perspectives:
\begin{itemize}
    \item \textbf{Humanities Perspective} (732 prompts): Covering cultural, historical, and geographical knowledge across 244 countries and regions, we construct three prompts per country, resulting in a total of 732 prompts related to history, culture, and region-specific contexts.
    \item \textbf{Nature Perspective} (340 prompts): Covering six scientific disciplines including Astronomy, Biology \& Medicine, Chemistry, Earth Sciences, Physics, and Cross-disciplinary Science.
\end{itemize}

Each prompt requires models not only to recall relevant factual knowledge but also to perform implicit reasoning to accurately instantiate details that are not explicitly mentioned but logically necessary.

\subsection{Dataset Curation Process}

\begin{figure}[htbp]
    \centering
    \includegraphics[width=\linewidth]{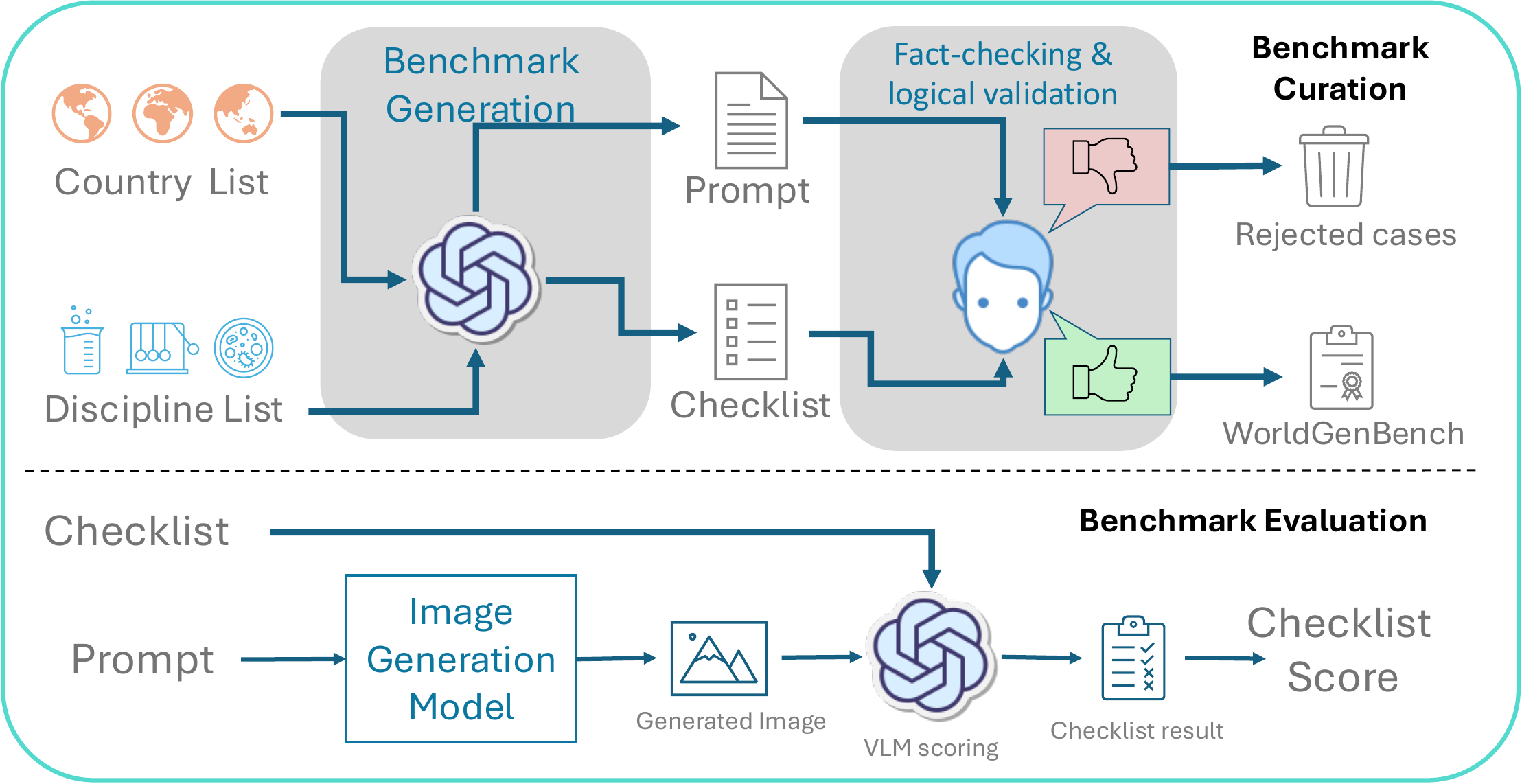}
    \caption{The construction and evaluation pipeline of WorldGenBench.}
\vspace{-5mm}
    \label{fig:cons}
\end{figure}

To ensure that each sample in WorldGenBench supports meaningful and rigorous evaluation of T2I models, as shown in Figure. \ref{fig:cons}, we adopt a multi-step curation pipeline that includes automated generation, domain-informed filtering, and fine-grained human review.

\definecolor{boxgray}{gray}{0.96}
\definecolor{humcolor}{rgb}{0.6, 0.2, 0.2}
\definecolor{natcolor}{rgb}{0.2, 0.3, 0.6}

\begin{figure*}[ht]
    \centering
    \footnotesize
    \begin{tcolorbox}[colback=boxgray, colframe=black, width=\textwidth, sharp corners=south, boxrule=0.4pt, title=\textbf{WorldGenBench Case Examples: Humanities and Nature}]
        \begin{minipage}[t]{0.48\textwidth}
            \textcolor{humcolor}{\textbf{Humanities Example — Afghanistan}} \\[0.5em]
            \textbf{\textcolor{blue}{Prompt (EN):}} \\
            \textcolor{black}{In the late fall of 2001, Nabi, a farmer in the Bamiyan Valley, stood at the edge of a terraced field and gazed at the remains of the Great Buddha in the distance.} \\[0.5em]
            \textbf{\textcolor{blue}{Visual Checkpoints (3 of 10 shown):}}
            \begin{itemize}
                \item \textbf{Bamiyan Buddhas' niches show blasting } — \textcolor{gray}{Reflects 2001 Taliban bombing}
                \item \textbf{Farmer wears Hazara traditional dress} — \textcolor{gray}{Region-specific cultural detail}
                \item \textbf{Terraces with traditional stone berm structure} — \textcolor{gray}{Typical erosion control in the valley}
                \item \textcolor{gray}{$\cdots$ 7 additional visual checkpoints omitted}
            \end{itemize}
        \end{minipage}
        \hfill
        \begin{minipage}[t]{0.48\textwidth}
            \textcolor{natcolor}{\textbf{Nature Example — Physics: EMF Curve}} \\[0.5em]
            \textbf{\textcolor{blue}{Prompt (EN):}} \\
            \textcolor{black}{Draw a scientific schematic showing a graph of induced EMF over time, using sinusoidal waveform.} \\[0.5em]
            \textbf{\textcolor{blue}{Visual Checkpoints (3 of 10 shown):}}
            \begin{itemize}
                \item \textbf{Maximum slope at zero-crossing points} — \textcolor{gray}{Indicates peak flux change rate}
                \item \textbf{Flat tangents at crests/troughs} — \textcolor{gray}{Flux change is zero at extrema}
                \item \textbf{Symmetry of positive and negative half cycles} — \textcolor{gray}{Follows electromagnetic induction laws}
                \item \textcolor{gray}{$\cdots$ 7 additional visual checkpoints omitted}
            \end{itemize}
        \end{minipage}
    \end{tcolorbox}
    \caption{
    Each \textbf{WorldGenBench} case includes a natural-language prompt and three key visual checkpoints (out of 10). 
    \textit{Left}: Humanities (historical grounding); 
    \textit{Right}: Nature (scientific schematic fidelity). 
    Within each checklist, the \textbf{bolded part} represents the expected \textbf{visual item}, while the \textcolor{gray}{gray text} provides the corresponding \textcolor{gray}{explanation or rationale}.
    }
    \label{fig:5}
\end{figure*}
\paragraph{Prompt Generation and Filtering.} 
Prompts are generated using GPT-4o to ensure breadth and topic relevance across diverse domains. For the Humanities Perspective, prompts are created for each country and region, highlighting culturally and historically significant scenes, figures, or events. For the Nature Perspective, prompts are distributed among core scientific disciplines, each requiring domain-specific factual grounding. After generation, all prompts undergo human verification to filter out instances that are overly generic, ambiguous, factually incorrect, and culturally inappropriate. After filtering, 1,072 prompts remain, comparable to WISE\cite{niu2025wise} (1,000) and larger than PhyBench\cite{meng2024phybench} (700).

\paragraph{Checklist Construction.} 
For each validated prompt, we first let GPT-4o construct a corresponding knowledge checklist, then perform human validation. Each checklist initially contains 10 candidate items, specifying concrete expectations for image content, such as a required object, scene layout, physical condition, or inferred event. As the necessity and completeness of evaluation criteria are unverifiable for open-ended T2I tasks, we prioritize necessary features rather than exhaustive descriptions. For factual grounding, checklist items directly reference world knowledge (e.g., traditional clothing styles, scientific phenomena). For implicit reasoning, checklist items capture logical consequences that are not overtly stated in the prompt but are semantically necessary (e.g., wet surfaces in a rainy scene, historically accurate equipment in a medieval setting).
Each checklist is accompanied by explanations clarifying the rationale behind each requirement, ensuring the evaluation is both interpretable and reproducible.

\paragraph{Human Annotation and Validation.} 
To ensure the quality of WorldGenBench, all prompts and their checklists are manually reviewed. Annotators verify that (i) the factual information in each prompt is historically and scientifically correct, and (ii) the checklist expectations follow logically from the prompt. Each checklist item is further examined to avoid redundancy and to guarantee that it corresponds to an independently visible element in generated images. After refinement, 9.65 items remain per prompt in average. Examples from both Humanities and Nature domains (Figure ~\ref{fig:5}) are included to illustrate typical prompt–checklist structures and how world knowledge and implicit reasoning are systematically encoded.

\subsection{Evaluation: Knowledge Checklist Score}

It is important to emphasize that our proposed "Knowledge Checklist Score" evaluation framework focuses on a core, yet previously overlooked, aspect. This framework assesses a model's ability to integrate world knowledge and perform implicit reasoning to ensure factual accuracy and logical coherence in image content.
As such, this evaluation method deliberately does not assess metrics such as aesthetic quality, creativity, or alignment with users' subjective preferences. While these subjective dimensions are crucial for practical applications, they are already addressed by other evaluation benchmarks. The goal of WorldGenBench is to provide a quantifiable and interpretable diagnostic tool for assessing a model’s deep semantic understanding capabilities. As shown in Figure~\ref{fig:5}, it focuses exclusively on assessing models' world knowledge and implicit reasoning capabilities. To this end, for each text-to-image prompt, we construct a corresponding checklist, where each item represents a specific attribute that we expect the T2I model to generate based on its internal knowledge and reasoning abilities. We then employ a state-of-the-art vision-language model, GPT-4o, to evaluate the generated images by determining the number of checklist items satisfied for each image. The Knowledge Checklist Score for an individual image is computed as the ratio of satisfied items to the total number of checklist items. The overall model performance is assessed by calculating the Average Knowledge Checklist Score across all generated images. The score is normalized to a range of 0 to 100.
\section{Main Results}

\begin{figure*}[h!]
    \centering
    \includegraphics[width=0.80\linewidth]{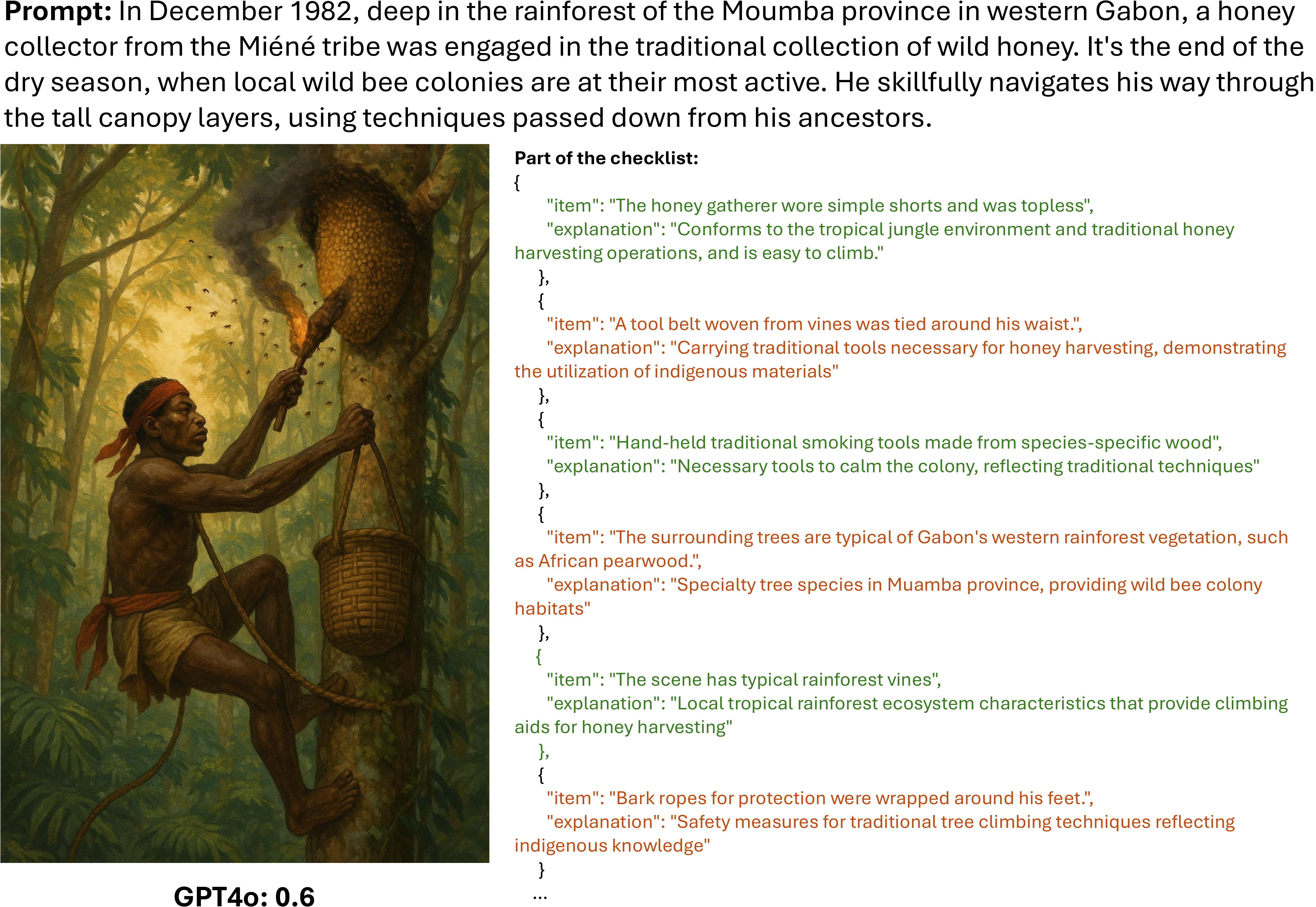}
    \caption{An example case from GPT4o. In the checklist, red text indicates that the model did not score on the corresponding item, while green text indicates that the model received a score. The red-text items underscore a critical gap in the model's capabilities: the discrepancy between generating images from general knowledge versus expert, domain-specific information. For instance, the model failed to render accurate botanical details ('African pearwood') and specific indigenous technologies ('Bark ropes'), demonstrating a limitation in factual and cultural fidelity, even while the overall scene appears plausible.}
    \label{fig:a3}
\end{figure*}

\definecolor{lightgray}{gray}{0.95}
\definecolor{lightblue}{RGB}{220,230,241}
\definecolor{lightgreen}{RGB}{220,240,220}
\definecolor{rowgray}{gray}{0.95}

\begin{table*}[h!]
    \centering
    \begin{minipage}{0.99\textwidth}
    \centering
    \resizebox{\linewidth}{!}{%
    \begin{tabular}{c|c|c|c|c|c|c|c|c}
    \toprule
    \rowcolor{headBlue}
    \multicolumn{1}{c}{} & \multicolumn{8}{c}{\color{black}{Humanities (GPT-4o / Gemini 2.5 Pro)} }\\
    \rowcolor{headBlue}
    \hline
    \diagbox{\color{black}Models}{\color{black}Continent} & \color{black}AF & \color{black}AN & \color{black}AS & \color{black}EU & \color{black}NA & \color{black}OC & \color{black}SA & \color{black}Avg. \\
    \hline


    \rowcolor{appleGreen}
    HiDream-l1-Full\cite{hidream}  & 16.61 / 13.62 & 13.61 / 19.17 & 17.96 / 18.24 & 18.28 / 14.32 & 14.39 / 10.58 & 16.53 / 11.7 & 13.79 / 13.08 & 16.68 / 14.39 \\
    \rowcolor{appleGreen}
    SDv3.5-Large\cite{esser2024scalingrectifiedflowtransformers} & 11.82 / 12.46 & 13.24 / 7.41 & 13.43 / 14.97 & 12.72 / 14.33 & 11.46 / 11.81 & 11.91 / 9.1 & 15.57 / 11.81 & 12.57 / 11.70 \\
    \rowcolor{appleGreen}
    SDv3.5-Medium\cite{esser2024scalingrectifiedflowtransformers} & 12.08 / 18.02 & 11.94 / 18.06 & 12.44 / 21.94 & 11.40 / 20.1 & 10.15 / 19.06 & 13.33 / 18.91 & 12.69 / 19.17 & 11.85 / 19.32 \\
    \rowcolor{appleGreen}
    Playground-v2.5\cite{li2024playground} & 12.03 / 11.49 & 10.10 / 11.76 & 13.27 / 12.05 & 11.91 / 10.17 & 10.35 / 10.88 & 9.68 / 11.1 & 13.16 / 11.12 & 11.83 / 11.22 \\
    \rowcolor{appleGreen}
    FLUX.1-schnell\cite{flux2024} & 11.31 / 9.96 & 8.61 / 11.21 & 13.52 / 11.18 & 11.79 / 11.72 & 10.84 / 8.98 & 12.38 / 9.18 & 12.96 / 8.86 & 12.00 / 10.16 \\
    \rowcolor{appleGreen}
    SDXL\cite{podell2023sdxlimprovinglatentdiffusion}  & 10.89 / 1.46 & 9.09 / 2.5 & 11.47 / 1.02 & 10.14 / 1.08 & 9.94 / 0.89 & 9.54 / 2.11 & 10.74 / 1.99 & 10.55 / 1.58 \\
    \rowcolor{appleGreen}
    PixArt-alpha\cite{chen2023pixartalpha} & 10.12 / 9.05 & 7.27 / 5.8 & 12.58 / 12.87 & 11.24 / 9.14 & 9.83 / 8.59 & 8.66 / 8.47 & 9.46 / 8.33 & 10.65 / 8.89 \\
    \rowcolor{appleGreen}
    FLUX.1-dev\cite{flux2024} & 8.43 / 6.46 & 11.30 / 4.44 & 10.15 / 9.51 & 10.59 / 7.58 & 8.23 / 7.13 & 8.43 / 6.37 & 9.63 / 8.89 & 9.36 / 7.20 \\
    
    \midrule
    \rowcolor{appleGray}
    Ming-Lite-Uni\cite{ai2025mingliteuniadvancementsunifiedarchitecture} & 12.30 / 16.51 & 12.68 / 20.56 & 16.53 / 20.74 & 14.87 / 18.05 & 13.15 / 15.69 & 12.64 / 15.0 & 14.35 / 18.13 & 14.13 / 17.81 \\
    \rowcolor{appleGray}
    T2I-R1\cite{jiang2025t2i} & 10.43 / 15.92 & 12.22 / 18.89 & 12.04 / 18.57 & 9.41 / 19.11 & 9.75 / 17.5 & 8.52 / 13.85 & 11.83 / 19.24 & 11.06 / 17.58 \\
    \rowcolor{appleGray}
    Show-o-512\cite{xie2024showo} & 10.99 / 8.01 & 8.38 / 4.17 & 12.60 / 10.01 & 11.91 / 8.06 & 12.13 / 7.93 & 8.98 / 6.14 & 15.34 / 11.53 & 11.75 / 7.98 \\
    \rowcolor{appleGray}
    Emu3\cite{wang2024Emu3} & 10.44 / 16.21 & 9.35 / 18.89 & 11.85 / 19.48 & 12.57 / 20.56 & 9.84 / 17.05 & 10.00 / 14.73 & 11.77 / 19.24 & 11.13 / 18.02 \\
    \rowcolor{appleGray}
    Harmon-1.5B\cite{wu2025harmon} & 10.04 / 13.55 & 7.69 / 11.85 & 10.86 / 17.36 & 9.56 / 16.08 & 8.99 / 14.79 & 9.05 / 16.14 & 12.17 / 14.87 & 9.96 / 14.95 \\
    \rowcolor{appleGray}
    SimpleAR(RL)\cite{wang2025simplear}& 7.40 / 12.74 & 4.75 / 9.54 & 8.13 / 13.94 & 8.91 / 13.71 & 7.13 / 12.15 & 7.61 / 13.34 & 8.92 / 13.73 & 7.90 / 12.74 \\
    \rowcolor{appleGray}
    GoT-6B\cite{fang2025got} & 7.89 / 7.3 & 9.26 / 8.61 & 9.17 / 8.81 & 8.03 / 6.58 & 7.83 / 7.26 & 7.07 / 8.7 & 7.49 / 8.77 & 8.12 / 8.00 \\
    \rowcolor{appleGray}
    SimpleAR(SFT)\cite{wang2025simplear}& 7.97 / 12.28 & 8.28 / 10.28 & 8.40 / 13.57 & 7.28 / 13.4 & 7.61 / 12.83 & 6.63 / 11.16 & 8.21 / 13.25 & 7.75 / 12.40 \\
    \rowcolor{appleGray}
    JanusPro-7B\cite{chen2025janus} & 6.87 / 12.57 & 5.45 / 10.37 & 8.57 / 15.41 & 9.22 / 16.71 & 5.28 / 11.69 & 5.24 / 10.91 & 8.46 / 15.93 & 7.41 / 13.37 \\
    \rowcolor{appleGray}
    VILA-u-7B-256\cite{wu2024vila} & 6.19 / 4.39 & 3.74 / 1.67 & 6.73 / 4.85 & 5.42 / 3.32 & 5.23 / 3.5 & 3.67 / 3.72 & 4.58 / 5.24 & 5.62 / 3.81 \\
    \rowcolor{appleGray}
    Lumina-mGPT-2.0\cite{lumina-mgpt-2.0} & 4.99 / 5.04 & 5.00 / 1.67 & 7.22 / 6.82 & 7.00 / 5.38 & 5.05 / 4.39 & 5.07 / 4.44 & 5.05 / 5.74 & 5.94 / 4.78 \\
    \rowcolor{appleGray}
    JanusFlow-1.3B\cite{chen2025janus} & 4.27 / 3.88 & 3.74 / 5.09 & 4.94 / 4.56 & 4.77 / 4.11 & 3.67 / 3.64 & 1.90 / 1.98 & 5.58 / 4.94 & 4.26 / 4.03 \\
    \rowcolor{appleGray}
    JanusPro-1B\cite{chen2025janus} & 3.34 / 7.53 & 5.93 / 7.69 & 4.02 / 8.93 & 2.97 / 8.37 & 2.07 / 5.9 & 3.60 / 6.87 & 5.33 / 8.39 & 3.41 / 7.67 \\
    
    \midrule
    

    \rowcolor{appleBlue}
    GPT-4o~\cite{gpt4o}  & 23.96 / 27.56 & 17.47 / 24.34 & 26.22 / 32.14 & 25.12 / 30.7 & 24.98 / 29.8 & 21.03 / 26.49 & 23.29 / 30.21 & 24.46 / 28.75 \\
    \rowcolor{appleBlue}
    Ideogram 2.0~\cite{ideogram} & 10.65 / 17.79 & 5.93 / 15.46 & 14.77 / 24.0 & 13.37 / 22.0 & 11.35 / 20.85 & 12.84 / 18.03 & 10.42 / 19.74 & 12.42 / 19.70 \\
    \rowcolor{appleBlue}
    Midjourney-v6~\cite{midj} & 11.62 / 22.49 & 9.01 / 24.56 & 13.21 / 28.16 & 14.16 / 29.23 & 11.34 / 26.58 & 10.23 / 23.09 & 12.05 / 23.97 & 12.33 / 25.44 \\
    \bottomrule
    \end{tabular}%
    }
    \caption{\textbf{Knowledge Checklist Score on the \textsc{Humanities} split.}
    The scores on the left and right of the slash (/) are evaluated by GPT-4o and Gemini 2.5 Pro, respectively. Columns report per-continent scores
    (\textbf{AF}=Africa, \textbf{AN}=Antarctica, \textbf{AS}=Asia,
    \textbf{EU}=Europe, \textbf{NA}=North America, \textbf{OC}=Oceania,
    \textbf{SA}=South America; \textbf{Avg.}=mean across continents).
    The \colorbox{appleGreen}{\strut upper block} lists \emph{diffusion} models,
    the \colorbox{appleGray}{\strut middle block} lists \emph{autoregressive}
    models, and the \colorbox{appleBlue}{\strut lower block} contains
    \emph{proprietary} systems.}
    \label{tab:1}
    \end{minipage}
\end{table*}



\begin{table*}[h!]
    \centering
    \resizebox{0.99\textwidth}{!}{%
    \begin{tabular}{c|c|c|c|c|c|c|c}
    \toprule
    \rowcolor{headBlue}
    \multicolumn{1}{c}{} & \multicolumn{7}{c}{\color{black}Nature (GPT-4o / Gemini 2.5 pro)} \\
    \rowcolor{headBlue}
    \hline
    \diagbox{\color{black}Models}{\color{black}Fields} & \color{black}ASTR & \color{black}BIO\&MED & \color{black}CHEM & \color{black}EASC & \color{black}PHYS & \color{black}XDIS & \color{black}Avg.  \\
    \hline
    
    \rowcolor{appleGreen}
    SDv3.5-Large\cite{esser2024scalingrectifiedflowtransformers} & 8.07 / 21.48 & 8.80 / 21.51 & 4.74 / 14.06 & 11.15 / 22.31 & 9.16 / 21.19 & 4.44 / 18.73 & 7.93 / 19.88  \\
    \rowcolor{appleGreen}
    FLUX.1-schnell\cite{flux2024} & 8.39 / 15.76 & 5.12 / 14.24 & 4.83 / 14.59 & 7.50 / 19.91 & 9.20 / 14.21 & 5.67 / 15.7 & 6.87 / 15.74 \\
    \rowcolor{appleGreen}
    HiDream-l1-Full~\cite{hidream} & 8.28 / 17.23 & 3.42 / 14.47 & 5.69 / 14.29 & 7.33 / 16.42 & 8.36 / 22.86 & 6.64 / 19.75 & 6.68 / 17.50  \\
    \rowcolor{appleGreen}
    FLUX.1-dev\cite{flux2024} & 5.15 / 9.12 & 4.08 / 8.76 & 4.33 / 9.46 & 5.41 / 15.46 & 7.50 / 15.34 & 3.02 / 13.67 & 5.19 / 11.97  \\
    \rowcolor{appleGreen}
    SDv3.5-Medium\cite{esser2024scalingrectifiedflowtransformers} & 3.33 / 11.6 & 2.58 / 11.91 & 2.07 / 9.85 & 6.87 / 22.64 & 3.83 / 10.43 & 4.84 / 15.8 & 4.06 / 13.70  \\
    \rowcolor{appleGreen}
    SDXL\cite{podell2023sdxlimprovinglatentdiffusion}  & 2.53 / 8.27 & 4.76 / 13.73 & 1.13 / 9.0 & 5.65 / 19.14 & 2.25 / 11.64 & 2.92 / 12.53 & 3.29 / 12.38  \\
    \rowcolor{appleGreen}
    PixArt-alpha\cite{chen2023pixartalpha} & 5.75 / 12.47 & 4.57 / 9.4 & 1.55 / 8.11 & 3.46 / 15.48 & 2.09 / 7.61 & 2.86 / 9.44 & 3.19 / 10.42  \\
    \rowcolor{appleGreen}
    Playground-v2.5\cite{li2024playground} & 5.66 / 15.46 & 3.27 / 7.4 & 1.23 / 7.0 & 4.27 / 13.49 & 2.33 / 7.93 & 2.27 / 8.87 & 3.07 / 10.03  \\
    \midrule
    \rowcolor{appleGray}
    T2I-R1\cite{jiang2025t2i} & 10.38 / 15.2 & 4.01 / 7.27 & 7.10 / 6.5 & 3.42 / 8.36 & 6.14 / 8.25 & 6.54 / 10.95 & 6.49 / 9.42 \\
    \rowcolor{appleGray}
    Ming-Lite-Uni\cite{ai2025mingliteuniadvancementsunifiedarchitecture} & 4.78 / 8.78 & 1.78 / 3.64 & 2.06 / 3.02 & 4.87 / 7.43 & 5.24 / 8.5 & 2.00 / 11.47 & 3.59 / 7.14 \\
    \rowcolor{appleGray}
    Show-o-512\cite{xie2024showo} & 7.01 / 11.11 & 1.48 / 4.64 & 2.78 / 8.34 & 3.39 / 8.91 & 5.03 / 10.13 & 3.34 / 8.11 & 3.76 / 8.54  \\
    \rowcolor{appleGray}
    JanusPro-7B\cite{chen2025janus} & 5.19 / 14.13 & 1.51 / 6.69 & 1.89 / 5.52 & 3.21 / 9.92 & 3.84 / 5.98 & 4.68 / 9.89 & 3.30 / 8.69  \\
    \rowcolor{appleGray}
    Harmon-1.5B\cite{wu2025harmon} & 5.20 / 9.8 & 5.25 / 3.6 & 2.82 / 5.33 & 2.99 / 9.6 & 1.64 / 5.75 & 1.67 / 6.24 & 3.15 / 6.72  \\
    \rowcolor{appleGray}
    Emu3\cite{wang2024Emu3} & 4.50 / 15.39 & 3.43 / 4.2 & 0.83 / 4.87 & 5.56 / 11.12 & 1.35 / 6.05 & 2.34 / 9.44 & 3.05 / 8.51 \\
    \rowcolor{appleGray}
    VILA-u-7B-256\cite{wu2024vila} & 1.56 / 2.86 & 1.25 / 0.2 & 1.15 / 1.86 & 4.39 / 2.76 & 3.28 / 2.46 & 1.84 / 1.84 & 2.46 / 2.00  \\
    \rowcolor{appleGray}
    SimpleAR(SFT)\cite{wang2025simplear}& 2.06 / 8.61 & 2.11 / 4.0 & 0.53 / 2.0 & 4.89 / 5.25 & 0.82 / 3.62 & 1.78 / 6.04 & 2.28 / 4.92  \\
    \rowcolor{appleGray}
    SimpleAR(RL)\cite{wang2025simplear}& 2.98 / 5.7  & 1.54 / 2.2 & 0.85 / 3.02 & 2.60 / 6.6 & 1.83 / 3.43 & 2.00 / 6.02 &  1.97 / 4.50 \\
    \rowcolor{appleGray}
    GoT-6B\cite{fang2025got} & 2.62 / 5.37 & 0.69 / 1.22 & 1.57 / 3.24 & 1.36 / 6.62 & 1.33 / 2.4 & 1.96 / 8.57 & 1.53 / 4.57 \\
    \rowcolor{appleGray}
    Lumina-mGPT-2.0\cite{lumina-mgpt-2.0} & 2.03 / 8.72 & 0.64 / 2.64 & 1.11 / 6.67 & 2.12 / 4.33 & 1.73 / 6.24 & 0.60 / 6.22 & 1.41 / 5.80 \\
    \rowcolor{appleGray}
    JanusPro-1B\cite{chen2025janus} & 1.81 / 6.6 & 0.00 / 1.0 & 0.48 / 3.02 & 0.70 / 2.46 & 1.23 / 2.1 & 1.28 / 6.49 & 0.91 / 3.61  \\
    \rowcolor{appleGray}
    JanusFlow-1.3B\cite{chen2025janus}& 0.57 / 3.33 & 0.00 / 0.61 & 1.67 / 0.86 & 0.34 / 4.04 & 0.43 / 2.4 & 0.94 / 2.5 & 0.60 / 2.29 \\
    
    \midrule
    \rowcolor{appleBlue}
    GPT-4o~\cite{gpt4o}  & 19.75 / 28.77 & 15.29 / 33.7 & 18.85 / 31.23 & 17.22 / 31.96 & 28.86 / 40.69 & 15.41 / 32.36 & 19.61 / 33.12  \\
    \rowcolor{appleBlue}
    Ideogram 2.0~\cite{ideogram}  & 11.15 / 23.03 & 7.14 / 11.42 & 8.33 / 12.72 & 7.63 / 21.76 & 12.08 / 14.88 & 9.82 / 21.81 & 9.34 / 17.60  \\
    \rowcolor{appleBlue}
    Midjourney-v6~\cite{midj} & 5.94 / 16.83 & 6.92 / 12.65 & 4.60 / 12.81 & 8.59 / 28.88 & 3.75 / 10.24 & 4.59 / 19.06 & 5.77 / 16.75  \\
    \bottomrule
    \end{tabular}%
    } 
    \caption{\textbf{Knowledge Checklist Score on the \textsc{Nature} split.}
    The scores on the left and right of the slash (/) are evaluated by GPT-4o and Gemini 2.5 Pro, respectively. Columns denote scientific disciplines
    (\textbf{ASTR}=Astronomy, \textbf{BIO\&MED}=Biology \& Medicine,
    \textbf{CHEM}=Chemistry, \textbf{EASC}=Earth Sciences,
    \textbf{PHYS}=Physics, \textbf{XDIS}=Cross-Disciplinary,
    \textbf{Avg.}=mean across disciplines).
    The \colorbox{appleGreen}{\strut upper block} lists \emph{diffusion} models,
    the \colorbox{appleGray}{\strut middle block} lists \emph{autoregressive} models,
    and the \colorbox{appleBlue}{\strut lower block} contains \emph{proprietary} systems.}

    \label{tab:2}
    \vspace{-6mm}
\end{table*}

\subsection{Experimental protocol.}
We report \textbf{Average Knowledge–Checklist Scores} on the \textsc{Humanities} and \textsc{Nature} splits for $24$ contemporary T2I systems: $8$ diffusion models, $13$ autoregressive (AR) models, and the $3$ leading proprietary systems. The judging models used were GPT-4o and Gemini 2.5 Pro. The open-source models were run on a server with 8*A800 80GB GPUs, using default-generation hyper-parameters are kept unchanged to isolate intrinsic knowledge–reasoning capacity. We have shown all the results in Table~\ref{tab:1} (Humanities) and Table~\ref{tab:2} (Nature).

\subsection{Quantitative Analysis}

\begin{figure*}[h!] 
    \centering 
    
    \includegraphics[width=0.9\textwidth]{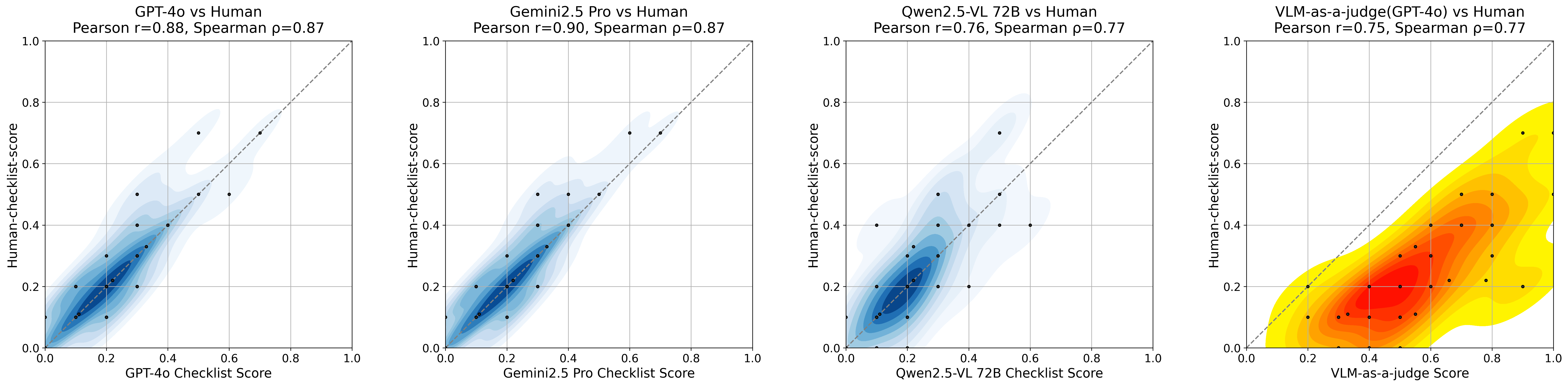} 
    

    \caption{Correlation on 50 sampled prompts between human checklist scores and automated scores from three VLMs. The high Pearson correlations for GPT-4o, Gemini 2.5 Pro, and Qwen2.5-VL 72B demonstrate strong human-VLM consistency, with most data points falling along the y=x line (Blue). In contrast, the VLM-as-a-judge method fails to align with the human checklist score, as its sample points are scattered and do not follow the diagonal (Red).}
    \vspace{-4mm}
    \label{fig:vlm} 
\end{figure*}

\begin{figure}[h!]
    \centering
    \includegraphics[width=0.9 \linewidth]{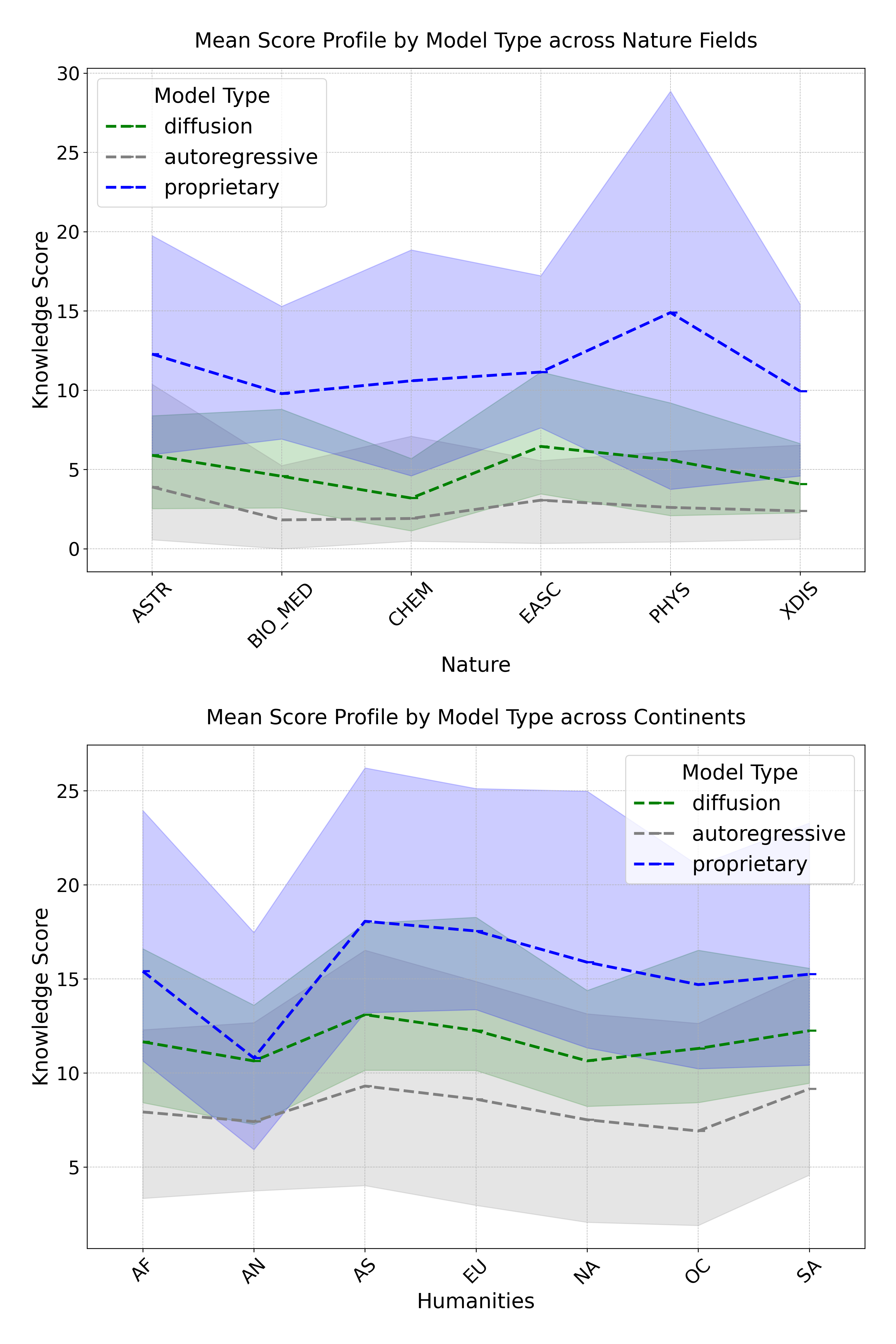}
    \caption{Mean Knowledge Score profiles for diffusion, autoregressive, and proprietary models on the WorldGenBench benchmark. The top chart compares performance across six natural science fields, and the bottom chart compares performance across seven continents for humanities. Shaded areas represent variance across models within each model type. Proprietary models consistently outperform diffusion and autoregressive models across all domains.}
    \vspace{-4mm}
    \label{fig:cons1}
\end{figure}

\paragraph{Correlation Analysis}
We employ \textsc{GPT-4o} and \textsc{Gemini 2.5 Pro} as judging models to evaluate a series of models under test on various tasks across Humanities and Nature. To validate the fairness of our evaluation framework, we calculate the \textbf{Pearson correlation coefficient} between the scoring results from the two judging models. First, for each model under test, we compute the correlation between the two sets of scores obtained across all test items, yielding an average correlation coefficient of \textbf{0.72} across all models. More critically, we directly compare the average scores assigned by the two judges to all the test models. The results show a correlation of \textbf{0.77} for the average scores in the Humanities split and an even higher correlation of \textbf{0.92} in the Nature split. These high positive correlations provide strong evidence that even though some evaluation data was generated by \textsc{GPT-4o}, its scoring standards as a judge remain highly consistent with those of \textsc{Gemini 2.5 Pro}. This indicates that our evaluation is fair and that concerns about potential "loop bias" are mitigated.

\paragraph{Domain difficulty gap.}
All models exhibit a pronounced drop when moving from culturally oriented prompts
to scientifically oriented ones.
The best open-source diffusion model ({\sc SD-v3.5-Large}) declines from
$12.57$ (\textsc{Humanities}) to $7.93$ (\textsc{Nature},    $\!\!\downarrow 37\%$),
while the strongest open-source AR model ({\sc Show-o-512}) falls from
$11.75$ to $3.76$ (   $\!\!\downarrow 68\%$).
The gap confirms that existing training corpora are substantially richer in
visual \emph{cultural} cues than in \emph{scientific} diagrams or physics-constrained scenes.

\paragraph{Architecture.}
On the open–source leaderboard, \textbf{diffusion} models dominate the \textsc{Nature} split, reaching $7.93$ (\textsc{SD-v3.5-Large}) versus only $3.76$ for the best \textbf{autoregressive} (AR) competitor (\textsc{Show-o-512}); in \textsc{Humanities} the gap nearly disappears.
This pattern indicates that the denoising prior of diffusion models is currently more adept at enforcing global spatial coherence and latent physical constraints, whereas sequence-only AR decoders remain brittle in scientific or physics-rich scenes.
However, \emph{GPT-4o} soars to $19.61$ on the same \textsc{Nature} split, more than doubling the best diffusion baseline.
Its success underscores that the AR paradigm is \emph{not} intrinsically limited; rather, today’s open-source AR models under-exploit its potential. Bridging this gap will require larger-scale knowledge conditioning, stronger image tokenizers, and reasoning-targeted training strategies to unlock GPT-4o-level performance in an open setting.

\paragraph{Geographic and disciplinary biases.}
A finer–grained break-down uncovers substantive imbalances hidden by aggregate means.
Across all diffusion backbones we observe a \emph{north–south} performance gradient: scores in \textit{AF} (Africa) and \textit{OC} (Oceania) trail those in \textit{EU} and \textit{AS} by up to \textbf{4.0} Checklist points, even for the strongest open model (\textsc{SD-v3.5-Large}).  The same pattern re-appears for proprietary systems, albeit with a smaller gap, suggesting that the root cause is less architectural than \emph{data availability}.  Existing web-scale image corpora are heavily tilted toward Euro-American and East-Asian visual culture; WorldGenBench exposes how this skew translates into brittle cultural grounding when the prompt references under-represented regions.
A similar disparity arises along \textsc{Nature} sub-fields.  The best diffusion model reaches \textbf{11.15} on Earth-science (\textit{EASC}) prompts—largely dominated by iconic landforms and meteorological scenes—but plunges to \textbf{4.74} on Chemistry (\textit{CHEM}), where success hinges on correctly rendering molecular schematics, reaction arrows, or glassware configurations.  In other words, “photolike” natural phenomena are already well covered by current pre-training, whereas highly stylized, diagrammatic, or symbolic conventions remain an open frontier.  Addressing these domain gaps will likely require \emph{targeted curation} of textbook–style figures and \emph{structure-aware} generation objectives that go beyond pixel fidelity.


\paragraph{Visual case.}
Figure~\ref{fig:a3} shows a bare‑chested honey hunter from the Miéné tribe climbing a vine‑covered rainforest tree in western Gabon, with an overlayed checklist that marks in green the scene details GPT‑4o correctly identified (e.g., indigenous tools, rainforest vines) and in red the ones it missed, leading to an overall score of 0.6. The mix of correct and missed detections illustrates that while GPT‑4o can grasp broad contextual cues in complex images, its fine‑grained visual reasoning still has notable gaps.

\paragraph{Error analysis.}

Despite achieving visually plausible generations in many cases, our analysis reveals three recurring categories of failure across models. \textbf{Semantic hallucination} occurs when the generated image captures the general atmosphere of the prompt but omits key factual or inferential elements required by the knowledge checklist. For instance, in the Gabon honey-gathering scene in Figure.~\ref{fig:a3} , GPT-4o failed to depict essential regional details such as African pearwood trees and indigenous climbing gear, despite producing a contextually coherent rainforest image. Secondly, we observe \textbf{cultural underrepresentation}, particularly for prompts rooted in under-resourced regions such as Africa and Oceania(Shown in Figure.~\ref{fig:cons1}). This is reflected in lower average checklist scores and the absence of culturally specific visual elements, highlighting training data imbalances in global cultural coverage. Lastly, models exhibit a tendency of \textbf{overfitting to photographic data}, performing relatively well on photo-like scenes but failing in structured scientific visualizations. As shown in the Appendix, in the acid-base titration prompt, most models struggled to render key schematic features—such as molecular insets, labeled inflection points, and pKa indicators—indicating poor generalization to symbolic and diagrammatic tasks. These findings underscore the limitations of current T2I models in semantic fidelity, cultural inclusiveness, and scientific abstraction, despite advances in visual realism.

\subsection{Human/Multi-Model Consistency Validation}
To verify the Knowledge Checklist Score, we compared human ratings averaged from 10 annotators on 50 randomly sampled prompts against checklist-based scores by three VLMs, which are respectively GPT-4o, Gemini 2.5 Pro, and Qwen2.5-VL 72B, and a traditional VLM-as-a-judge baseline score. As shown in Figure~\ref{fig:vlm}, our checklist-based VLMs strongly align with human judgments, achieving Pearson/Spearman correlations of 0.88/0.87 (GPT-4o), 0.90/0.87 (Gemini), and 0.76/0.77 (Qwen2.5-VL 72B). Although Gemini demonstrates slightly higher consistency, we default to GPT-4o to maintain compatibility with prior benchmarks. Conversely, VLM-as-a-judge fails to align with human preference. Unlike VLM-as-a-judge, which is prone to associative bias, our Checklist Score enforces explicit, verifiable evidence, providing an interpretable, fine-grained diagnostic assessment of T2I models' factual grounding and reasoning. More data is available in Appendix.

\section{Conclusion}


The integration of world knowledge and implicit reasoning is essential for real-world text-to-image generation. Knowledge without reasoning produces shallow literalism; reasoning without knowledge leads to plausible but inaccurate outputs. Only through their synergy can models truly understand prompts and generate semantically coherent, context-aware images.
To evaluate this synergy, we introduced WorldGenBench, a benchmark that jointly assesses factual grounding and inferential capability across both cultural and scientific domains. Unlike prior benchmarks, it emphasizes fine-grained, checklist-based evaluation to capture not just what models render—but why.
Our evaluation of 24 state-of-the-art models shows that despite advances in visual quality, most systems fall short in combining knowledge and reasoning effectively—particularly in underrepresented regions and abstract scientific scenarios. Proprietary models like GPT-4o demonstrate the promise of such integration, while open-source models still face significant gaps.
WorldGenBench provides not only a diagnostic tool but a roadmap for building T2I systems that reason with what they know—and know what they reason about.

\clearpage
\bibliographystyle{ACM-Reference-Format}
\bibliography{main}

\clearpage
\section{Supplementary materials}

\begin{figure*}[t!]
    \centering
    \includegraphics[width=1\linewidth]{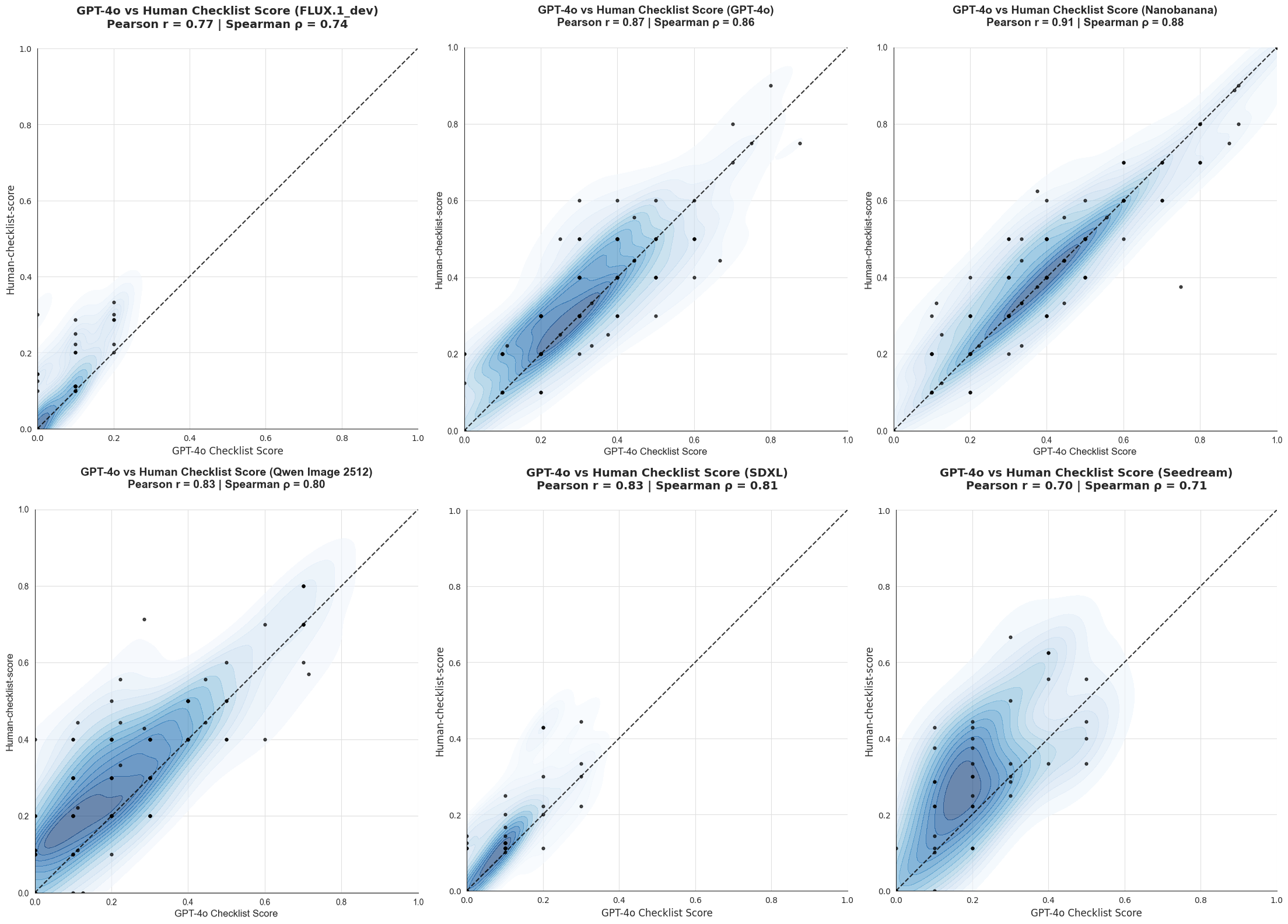}
    \caption{Correlation of GPT-4o and human evaluations. Results across six text-to-image models show strong alignment (Pearson $r$ from 0.70 to 0.91), demonstrating the evaluation's reliability against loop bias across different model families.}
    \label{fig:human_align}
\end{figure*}

\begin{table*}[t]
    \centering
    \scriptsize
    \begin{minipage}{0.99\textwidth}
    \centering
    \resizebox{\linewidth}{!}{%
    \begin{tabular}{c|c|c|c|c|c|c|c|c}
    \toprule
    \rowcolor{headBlue}
    \multicolumn{1}{c}{} & \multicolumn{8}{c}{\color{black}{Humanities}} \\
    \rowcolor{headBlue}
    \hline
    \diagbox{\color{black}Models}{\color{black}Continent} & \color{black}AF & \color{black}AN & \color{black}AS & \color{black}EU & \color{black}NA & \color{black}OC & \color{black}SA & \color{black}Avg. \\
    \hline

    \rowcolor{appleGreen}
    \textbf{Qwen Image} & 17.01 & 16.39 & 25.03 & 24.47 & 18.95 & 15.87 & 18.68 & 20.66 \\
    \rowcolor{appleGreen}
    \textbf{Qwen Image 2512} & 22.01 & 17.96 & 26.78 & 28.44 & 23.91 & 20.88 & 23.07 & 24.60 \\
    \rowcolor{appleGreen}
    \textbf{Z-Image} & 19.35 & 15.28 & 22.89 & 23.93 & 21.49 & 19.81 & 22.54 & 21.61 \\
    \rowcolor{appleGreen}
    \textbf{FLUX.2-dev} & 25.40 & 28.80 & 30.29 & 31.23 & 28.55 & 24.65 & 25.72 & 28.21 \\
    \rowcolor{appleGreen}
    \textbf{FLUX.2-klein} & 24.18 & 23.80 & 28.40 & 27.57 & 26.07 & 25.62 & 26.41 & 26.42 \\

    \midrule

    \rowcolor{appleGray}
    \textbf{GLM-Image} & 18.12 & 15.46 & 20.91 & 21.90 & 20.15 & 15.38 & 15.92 & 19.40 \\
    \rowcolor{appleGray}
    \textbf{BitDance} & 20.98 & 15.37 & 25.79 & 24.86 & 23.79 & 21.67 & 24.40 & 23.50 \\

    \midrule

    \rowcolor{appleBlue}
    \textbf{SeedDream} & 24.89 & 15.76 & 32.30 & 30.61 & 29.35 & 24.80 & 24.33 & 28.32 \\
    \rowcolor{appleBlue}
    \textbf{Nano Banana} & 32.76 & 29.07 & 36.63 & 37.02 & 34.66 & 33.44 & 34.87 & 34.96 \\

    \bottomrule
    \end{tabular}%
    }
    \caption{\textbf{Performance on the \textsc{Humanities} split.}
    Columns report per-continent scores
    (\textbf{AF}=Africa, \textbf{AN}=Antarctica, \textbf{AS}=Asia,
    \textbf{EU}=Europe, \textbf{NA}=North America, \textbf{OC}=Oceania,
    \textbf{SA}=South America; \textbf{Avg.}=mean across continents).
    The \colorbox{appleGreen}{\strut upper block} lists \emph{diffusion-based} models,
    the \colorbox{appleGray}{\strut middle block} lists \emph{AR-based} models,
    and the \colorbox{appleBlue}{\strut lower block} contains
    \emph{proprietary} systems.}
    \label{tab:upd_human}
    \end{minipage}
\end{table*}

\begin{table*}[t]
    \centering
    \scriptsize
    \begin{minipage}{0.99\textwidth}
    \centering
    \resizebox{\linewidth}{!}{%
    \begin{tabular}{c|c|c|c|c|c|c|c}
    \toprule
    \rowcolor{headBlue}
    \multicolumn{1}{c}{} & \multicolumn{7}{c}{\color{black}{Nature}} \\
    \rowcolor{headBlue}
    \hline
    \diagbox{\color{black}Models}{\color{black}Fields} & \color{black}ASTR & \color{black}BIO & \color{black}CHEM & \color{black}EASC & \color{black}PHYS & \color{black}XDIS & \color{black}Avg. \\
    \hline

    \rowcolor{appleGreen}
    \textbf{Qwen Image} & 26.97 & 23.00 & 20.48 & 36.45 & 34.11 & 13.92 & 27.73 \\
    \rowcolor{appleGreen}
    \textbf{Qwen Image 2512} & 28.20 & 23.38 & 24.50 & 36.54 & 38.24 & 25.28 & 30.49 \\
    \rowcolor{appleGreen}
    \textbf{Z-Image} & 17.73 & 10.42 & 13.93 & 28.51 & 23.29 & 7.50 & 18.56 \\
    \rowcolor{appleGreen}
    \textbf{FLUX.2-dev} & 41.69 & 32.87 & 32.20 & 49.81 & 46.05 & 38.33 & 40.72 \\
    \rowcolor{appleGreen}
    \textbf{FLUX.2-klein} & 29.50 & 20.88 & 23.26 & 43.55 & 35.46 & 20.36 & 30.59 \\

    \midrule

    \rowcolor{appleGray}
    \textbf{GLM-Image} & 21.06 & 12.53 & 19.40 & 19.38 & 29.98 & 10.12 & 20.32 \\
    \rowcolor{appleGray}
    \textbf{BitDance} & 27.18 & 15.79 & 16.35 & 36.51 & 26.17 & 19.42 & 24.21 \\

    \midrule

    \rowcolor{appleBlue}
    \textbf{SeedDream} & 26.84 & 20.62 & 18.55 & 41.56 & 25.16 & 18.01 & 26.39 \\
    \rowcolor{appleBlue}
    \textbf{Nano Banana} & 52.99 & 42.19 & 48.07 & 53.14 & 64.75 & 49.29 & 52.74 \\

    \bottomrule
    \end{tabular}%
    }
    \caption{\textbf{Performance on the \textsc{Nature} split.}
    Columns denote scientific disciplines
    (\textbf{ASTR}=Astronomy, \textbf{BIO}=Biology,
    \textbf{CHEM}=Chemistry, \textbf{EASC}=Earth Sciences,
    \textbf{PHYS}=Physics, \textbf{XDIS}=Cross-Disciplinary,
    \textbf{Avg.}=mean across disciplines).
    The \colorbox{appleGreen}{\strut upper block} lists \emph{diffusion-based} models,
    the \colorbox{appleGray}{\strut middle block} lists \emph{AR-based} models,
    and the \colorbox{appleBlue}{\strut lower block} contains
    \emph{proprietary} systems.}
    \label{tab:upd_nat}
    \end{minipage}
    \vspace{-6mm}
\end{table*}
\begin{table*}[t]
\centering
\scriptsize
\setlength{\tabcolsep}{7pt}
\renewcommand{\arraystretch}{1.0}
\resizebox{0.72\textwidth}{!}{%
\begin{tabular}{lccc}
\toprule
Model & Aesthetic & ImageReward & PickScore \\
\midrule
GoT & 0.9537 & 0.6343 & 0.7947 \\
JanusFlow1.3B & 0.8827 & 0.3800 & 0.7432 \\
Lumina-gpt2 & 0.9362 & -0.3082 & 0.3778 \\
Ming & 0.9764 & 0.7692 & 0.8669 \\
SDXL & 0.9729 & -0.1842 & 0.9211 \\
T2I-R1 & 0.7854 & 0.8011 & 0.7318 \\
emu3 & 0.9672 & 0.9097 & 0.8406 \\
fluxdev & 0.8778 & -0.6533 & -0.0355 \\
fluxsch & 0.8427 & 0.5828 & 0.6944 \\
gpt4o & 0.4680 & 0.1395 & 0.0534 \\
harmon & 0.9279 & 0.7257 & 0.8608 \\
hidream\_l1\_full & 0.9467 & 0.1080 & 0.7449 \\
ideogramv2 & 0.4289 & 0.3164 & 0.4679 \\
janus\_pro\_1b & 0.8737 & -0.3336 & 0.6424 \\
janus\_pro\_7b & 0.8752 & 0.2672 & 0.8646 \\
midjourney & 0.9589 & -0.8138 & 0.6096 \\
pgv25 & 0.9738 & 0.5827 & 0.8950 \\
pixart & 0.9346 & 0.7150 & 0.8940 \\
sd35l & 0.8885 & 0.8671 & 0.8145 \\
sd35m & 0.9498 & 0.7534 & 0.8943 \\
simple\_rl & 0.9376 & 0.1408 & 0.4636 \\
simple\_sft & 0.9559 & -0.1430 & 0.2598 \\
vila-u & 0.7599 & 0.7425 & 0.7115 \\
\midrule
\textbf{TOTAL} & \textbf{0.7322} & \textbf{0.5299} & \textbf{0.7089} \\
\bottomrule
\end{tabular}%
}
\caption{Pearson correlation coefficients between WorldGenBench scores and three traditional metrics (Aesthetic Score, ImageReward, and PickScore) across all evaluated models.}
\label{tab:metrics_correlation}
\vspace{-2mm}
\end{table*}

\subsection{Extended Loop Bias Validation}

To further examine potential loop bias, we extend the consistency analysis to six representative text-to-image models with varying performance levels, as shown in Figure~\ref{fig:human_align}. For each model, we compare GPT-4o checklist scores with human-annotated scores and analyze their correlations. Strong positive correlations are consistently observed across all models (Pearson $r$ ranging from 0.70 to 0.91). The alignment holds not only on average but across the full score range, with data points closely following the diagonal trend.

These results suggest that the agreement between GPT-4o-based evaluation and human judgment is stable across different model families and capability levels. In particular, the consistency persists for both weaker models and high-performing systems, indicating that the evaluation is not driven by model-specific biases.

Overall, the extended analysis provides additional evidence that the checklist-based evaluation reliably reflects human judgment, mitigating concerns about loop bias in the evaluation pipeline.

\subsection{Updated Evaluation and Discussion}
\textbf{Overview.}
We update the evaluation results using a more recent set of representative text-to-image models, as shown in Table~\ref{tab:upd_nat} and Table~\ref{tab:upd_human}. The overall trends remain broadly consistent with those reported in the main paper, and reflect the rapid evolution of model capabilities.

\textbf{Overall ranking.}
Although proprietary systems maintain their dominance on the benchmark, the open-source community is making rapid strides. Current open-source counterparts have successfully outperformed legacy proprietary systems, yet a performance gap remains when compared to the newest proprietary models like \textit{Nano Banana}. 

\textbf{Architecture comparison.}
While autoregressive (AR) models, such as BitDance and GLM-Image, have demonstrated measurable progress and slightly narrowed the performance gap with diffusion-based models, they still lag behind overall. Diffusion architectures continue to maintain a clear advantage across both the Humanities and Nature domains. This indicates that while AR models are improving, architectural differences still play a significant role in determining final performance.

\textbf{Bias and fine-grained patterns.}
Consistent with our original analysis, fine-grained biases persist across both geographic regions and scientific disciplines. In the Humanities split, regions such as Africa and Oceania continue to exhibit relatively lower scores, reflecting ongoing imbalances in cultural knowledge coverage. In the Nature split, chemistry-related tasks remain particularly challenging, while visually grounded domains such as Earth Sciences are comparatively better handled. These patterns indicate that limitations in both data distribution and reasoning over symbolic or structured knowledge remain key bottlenecks.

\textbf{Benchmark implications.} 
Beyond updating current model rankings, these results underscore the diagnostic utility of \textsc{WorldGenBench}. As capabilities evolve, the benchmark captures nuanced developmental trends, such as the marginal narrowing of the gap between autoregressive and diffusion models. 

However, the benchmark concurrently exposes critical, unresolved challenges preventing overall performance saturation, as even state-of-the-art systems fail to maintain consistent alignment across diverse tasks and exhibit persistent vulnerabilities in chemistry-related generation and the representation of culturally marginalized regions, thereby revealing that implicit reasoning and deep world-knowledge grounding are still fundamentally bottlenecked.

Taken together, these findings affirm \textsc{WorldGenBench} as a dynamic, longitudinal diagnostic framework rather than a static leaderboard. By simultaneously tracking progress and illuminating resilient flaws, it maintains its vital role in steering future research toward more robust, knowledge-intensive text-to-image systems.

\textbf{Benchmark implications.} 
The updated results validate \textsc{WorldGenBench} as more than a static evaluation metric; it serves as a longitudinal diagnostic framework. As models evolve, the benchmark effectively tracks progressive milestones

Despite this progress, the benchmark highlights that we are far from performance saturation. Even the most advanced proprietary systems exhibit critical failure modes. Persistent deficits in complex domains like chemistry and culturally underrepresented geographies demonstrate that current models still lack comprehensive world-knowledge grounding.

By simultaneously mapping capability advancements and exposing these resilient bottlenecks, \textsc{WorldGenBench} provides a clear, ongoing roadmap. It remains highly relevant for guiding the broader research community toward next-generation text-to-image systems driven by explicit knowledge and robust implicit reasoning.

\subsection{Relationship with Traditional Metrics}

To situate \textsc{WorldGenBench} within the broader evaluation landscape, we analyze its Pearson correlation with three established automated metrics, as detailed in Table~\ref{tab:metrics_correlation}.

First, the \textbf{Aesthetic Score} exhibits a robust overall correlation of $0.73$, serving as a sanity check that capable models generally maintain high visual fidelity. However, a closer examination reveals a telling decoupling on models renowned for high semantic intelligence, such as GPT-4o ($0.46$) and Ideogram ($0.42$). This divergence underscores a critical, often overlooked distinction: ``looking good'' is not synonymous with ``being correct.'' While traditional aesthetic metrics fixate on surface-level elements, \textsc{WorldGenBench} penetrates deeper to evaluate the structural integrity of logic and factual grounding.

Second, the limitations of legacy alignment metrics become even more pronounced. \textbf{PickScore} remains effective for evaluating traditional architectures (e.g., SDXL: $0.92$) but collapses to near-zero predictive power for state-of-the-art systems like Flux-dev ($-0.03$) and GPT-4o ($0.05$). More radically, \textbf{ImageReward} shows a strong \textit{negative correlation} with top-tier visual models (e.g., Midjourney: $-0.81$; Flux-dev: $-0.65$). These findings indicate that metrics like PickScore and ImageReward capture only the coarsest level of image-text alignment—mere associative matching. They fail to distinguish between a visually plausible hallucination and a rigorously accurate depiction. By exposing these blind spots, \textsc{WorldGenBench} fills a critical void, shifting the evaluation paradigm from superficial alignment to genuine \textit{real-world readiness}.

\begin{figure}[h]
    \centering
    \includegraphics[width=1\linewidth]{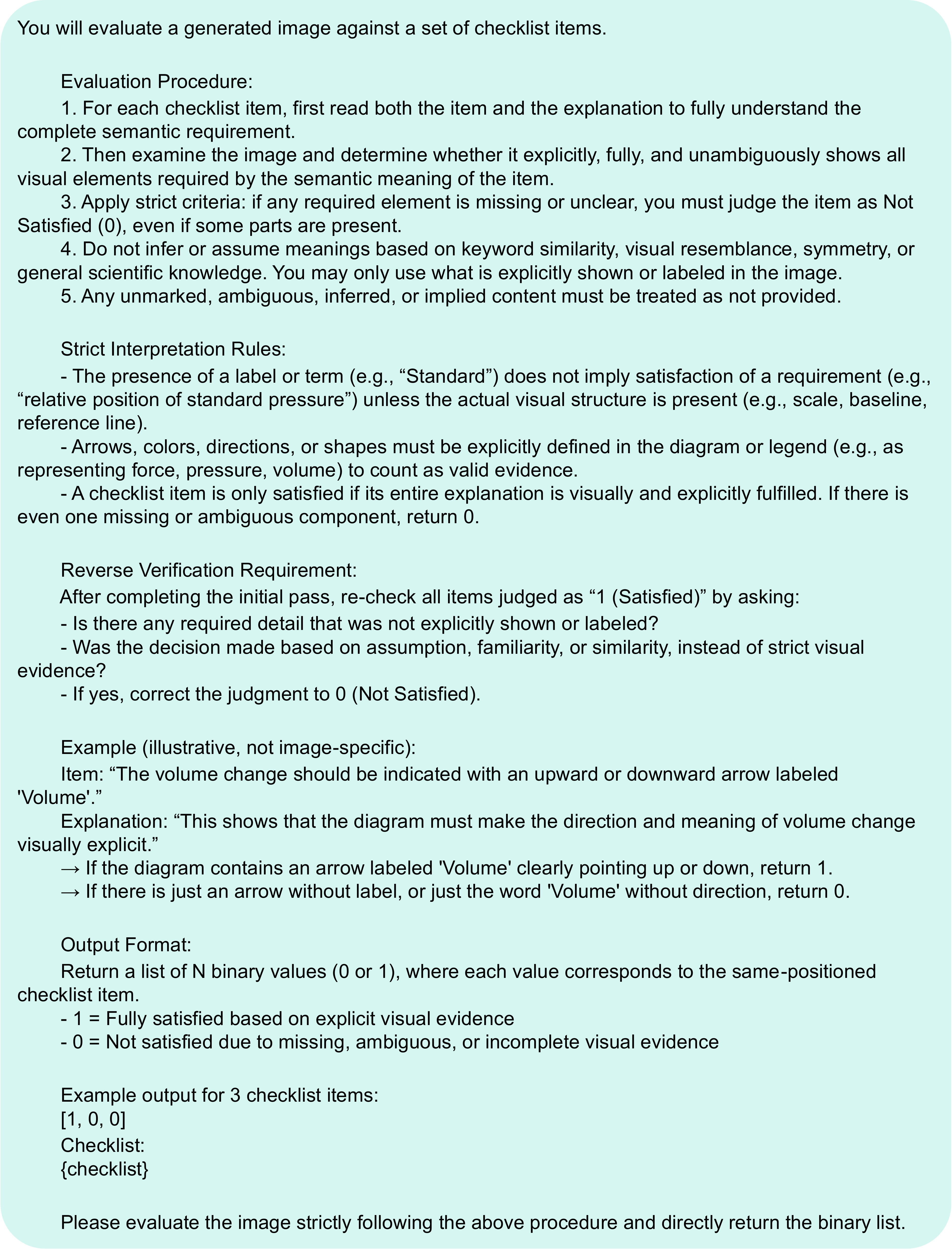}
    \caption{Prompt used for benchmarking with the default model, GPT-4o}
    \label{fig:prompt_demo}
\end{figure}

\section{Visual Cases}

We present two cases to compare models and one case to demonstrate the model's success and failure.

\begin{figure*}[h]
    \centering
    \includegraphics[width=0.85\linewidth]{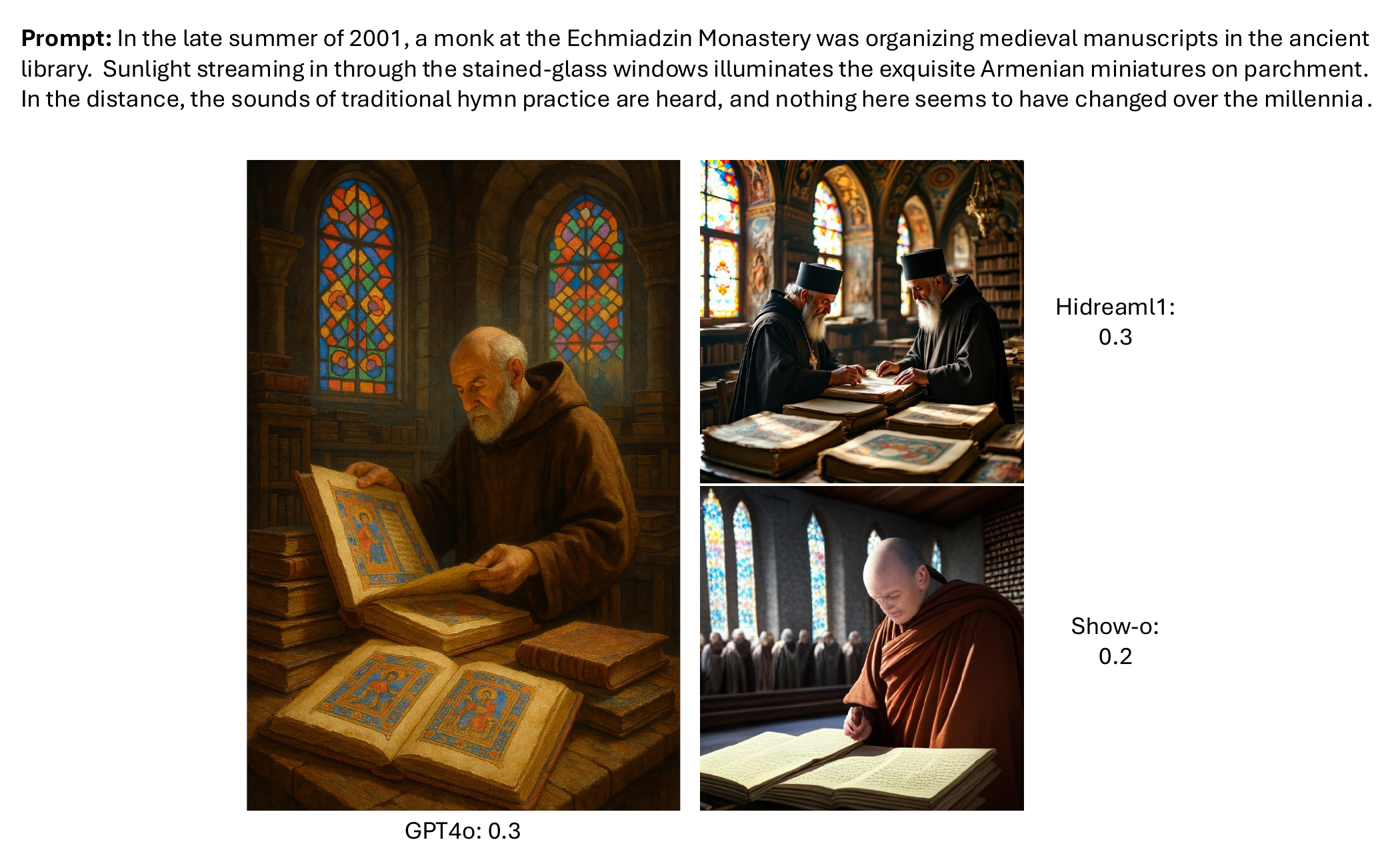}
    \caption{Visual Case 1: An example case from the humanities domain, including the prompt and the results from GPT-4o, HiDreaml1-Full, and Show-o. Checklist is in ~\ref{111}}
    \label{fig:a1}
\end{figure*}

\begin{figure*}
    \centering
    \includegraphics[width=0.85\linewidth]{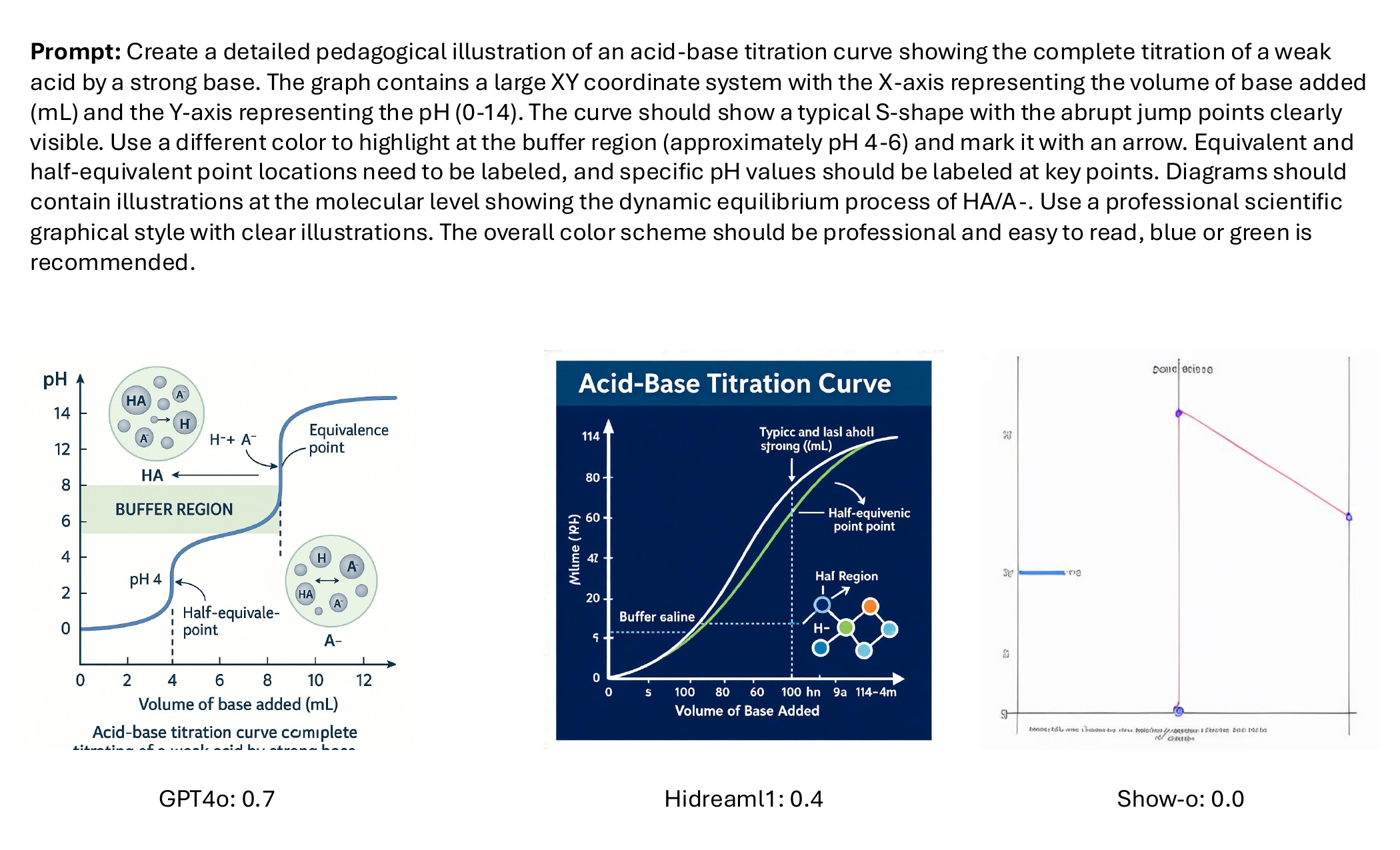}
    \caption{Visual Case 2: An example case from the Nature domain, including the prompt and the results from GPT-4o, HiDreaml1-Full, and Show-o. Checklist is in ~\ref{222}}
    \label{fig:a2}
\end{figure*}

The cases for model comparison are shown in Figure~\ref{fig:a1} and Figure~\ref{fig:a2}, while the case illustrating the detailed checklist scores is presented in Figure~\ref{fig:a3} in the main text.

\subsection{Checklist for Visual Case 1}
\label{111}

\textbf{Item:} The picture should feature typical Armenian church architecture \\
\textbf{Explanation:} The Echmiadzin Monastery has a unique architectural style, including conical domes and stone decorations.

\vspace{1em}
\textbf{Item:} Traditional style wooden bookshelves should appear \\
\textbf{Explanation:} As an ancient monastic library, it has retained its historic and traditional furnishings.

\vspace{1em}
\textbf{Item:} The friar shall wear the black traditional robe \\
\textbf{Explanation:} The monks of the Armenian Apostolic Church have their own specific traditional dress code.

\vspace{1em}
\textbf{Item:} There should be specialized tools on the desktop for restoration \\
\textbf{Explanation:} Antiquarian book restoration requires specific specialized tools, which are necessary working instruments.

\vspace{1em}
\textbf{Item:} Windows should present a characteristic painted pattern \\
\textbf{Explanation:} Stained glass in churches often contains religious themes and traditional motifs.

\vspace{1em}
\textbf{Item:} Manuscripts should have typical Armenian text \\
\textbf{Explanation:} Ancient Armenian manuscripts use their unique alphabet system.

\vspace{1em}
\textbf{Item:} There should be incense burners and candles in the room \\
\textbf{Explanation:} These are the traditional liturgical items of the Armenian Church, which are used year-round.

\vspace{1em}
\textbf{Item:} The walls should have old frescoes \\
\textbf{Explanation:} Monastery interiors often preserve historic religious frescoes.

\vspace{1em}
\textbf{Item:} The light should create a specific projection \\
\textbf{Explanation:} The text mentions sunlight filtering through the colored windows, a light effect that is one of the features of the church building.

\vspace{1em}
\textbf{Item:} Stone floors should be featured \\
\textbf{Explanation:} Monastery buildings are paved with local stone, reflecting architectural tradition.

\subsection{Checklist for Visual Case 2}

\label{222}

\textbf{Item:} The curve shows a smoother slope in the buffer region \\
\textbf{Explanation:} The buffer zone is resistant to pH changes, so the curve changes slowly in this region.

\vspace{1em}
\textbf{Item:} The curve shows a steep change near the equivalence point \\
\textbf{Explanation:} When the equivalence point is reached, the system is extremely sensitive to pH changes, and small additions of reagents can lead to dramatic pH changes.

\vspace{1em}
\textbf{Item:} The pH at the half-equivalent point should be equal to the pKa of the weak acid \\
\textbf{Explanation:} According to the Henderson-Hasselbalch equation, when [HA] = [A-], pH = pKa.

\vspace{1em}
\textbf{Item:} There is a significant difference between the slopes of the curves on either side of the buffer region \\
\textbf{Explanation:} Reflecting the difference in sensitivity of the system to pH changes at different stages.

\vspace{1em}
\textbf{Item:} Proportion of HA and A\textsuperscript{-} in molecular level insets versus curve \\
\textbf{Explanation:} The concentration of acid ions gradually increases and the concentration of un-ionized acid molecules decreases during the titration process.

\vspace{1em}
\textbf{Item:} Coordinate axis scales are evenly spaced and clear \\
\textbf{Explanation:} Specialized scientific charts require accurate data representation.

\vspace{1em}
\textbf{Item:} Buffer labeling should point to the inflection area of the curve \\
\textbf{Explanation:} The region of most significant buffering effect is near the half-equivalent point.

\vspace{1em}
\textbf{Item:} Legends contain math formulas or chemical equations \\
\textbf{Explanation:} Professional titration curves usually need to show the relevant theoretical basis.

\vspace{1em}
\textbf{Item:} The pH value at the equivalent point should be greater than 7 \\
\textbf{Explanation:} Weak acids react with strong bases and show basicity at the equivalence point.

\vspace{1em}
\textbf{Item:} The pH at the start of the curve should be close to the pH of the weak acid \\
\textbf{Explanation:} The initial state of the titration reflects the acidic character of the original solution.

\subsection{Evaluation Prompt for Knowledge Checklist Score}
We use GPT-4o to make the evaluation. The evaluation prompt is shown in Figure~\ref{fig:prompt_demo}.

\subsection{Models' average performance on countries}
The Table~\ref{tab:t1} shows the average performance of each Model on part of our countries ans regions. The name of models are shortened in the table, the following is som clarification:
The following describes the correspondence between model abbreviations and their complete names:

\texttt{JF1.3B} denotes \textit{JanusFlow1\_3B}, while \texttt{JP1B} and \texttt{JP7B} represent \textit{JanusPro1B} and \textit{JanusPro7B} respectively. The abbreviation \texttt{Lu} stands for \textit{Lumina-gpt2}, and \texttt{fdev}/\texttt{fsch} refer to \textit{fluxdev} and \textit{fluxsch}. The term \texttt{got} is used as-is without expansion.

The shorthand \texttt{4o} corresponds to OpenAI's \textbf{GPT-4o} model. \texttt{hm} indicates \textit{harmon}, and \texttt{hdl1} represents \textit{hidreaml1\_full}. For image generation models, \texttt{igv2} is \textit{ideogramv2}, \texttt{mj} means \textit{midjourney}

\texttt{pg} abbreviates \textit{pgv25}, and \texttt{px} stands for \textit{pixart}. The Stable Diffusion variants are denoted by \texttt{sd35l}, \texttt{sd35m}, and \texttt{sdxl} (retaining their exact names). \texttt{showo} refers to \textit{show-o}, while \texttt{sarrl} and \texttt{sarsft} represent \textit{simplear\_rl} and \textit{simplear\_sft} respectively. Finally, \texttt{vl} is short for \textit{vila-u}.

\begin{table*}[t]
\centering
\setlength{\tabcolsep}{1.5pt} 
\resizebox{\textwidth}{!}{%
\begin{tabular}{l *{21}{c}} 
\toprule
Class / Model & JF1.3B & JP1B & JP7B & Lu & fdev & fsch & got & 4o & hm & hdl1 & igv2 & mj & pg & px & sd35l & sd35m & sdxl & showo & sarrl & sarsft & vl \\ 
\midrule
Algeria & 0.00  & 0.03  & 0.00  & 0.13  & 0.03  & 0.20  & 0.07  & 0.47  & 0.17  & 0.17  & 0.23  & 0.03  & 0.23  & 0.17  & 0.17  & 0.07  & 0.03  & 0.20  & 0.03  & 0.07  & 0.10 \\ 
Angola & 0.00  & 0.00  & 0.00  & 0.07  & 0.07  & 0.05  & 0.00  & 0.13  & 0.00  & 0.05  & 0.00  & 0.05  & 0.15  & 0.00  & 0.00  & 0.05  & 0.10  & 0.10  & 0.05  & 0.00  & 0.00 \\ 
Benin & 0.03  & 0.00  & 0.03  & 0.07  & 0.00  & 0.10  & 0.07  & 0.07  & 0.05  & 0.13  & 0.13  & 0.10  & 0.05  & 0.00  & 0.10  & 0.05  & 0.10  & 0.07  & 0.07  & 0.07  & 0.00 \\ 
Botswana & 0.03  & 0.07  & 0.07  & 0.00  & 0.17  & 0.23  & 0.13  & 0.30  & 0.15  & 0.20  & 0.03  & 0.30  & 0.30  & 0.00  & 0.17  & 0.13  & 0.10  & 0.13  & 0.13  & 0.10  & 0.10 \\ 
Burkina Faso & 0.00  & 0.00  & 0.03  & 0.05  & 0.10  & 0.10  & 0.10  & 0.30  & 0.10  & 0.13  & 0.00  & 0.30  & 0.10  & 0.20  & 0.07  & 0.07  & 0.10  & 0.03  & 0.03  & 0.03  & 0.00 \\ 
Burundi & 0.00  & 0.00  & 0.10  & 0.00  & 0.13  & 0.17  & 0.13  & 0.20  & 0.17  & 0.27  & 0.07  & 0.25  & 0.17  & 0.10  & 0.10  & 0.20  & 0.17  & 0.20  & 0.10  & 0.10  & 0.00 \\ 
Cameroon & 0.03  & 0.07  & 0.20  & 0.00  & 0.15  & 0.17  & 0.23  & 0.40  & 0.27  & 0.27  & 0.27  & 0.10  & 0.23  & 0.30  & 0.23  & 0.37  & 0.17  & 0.10  & 0.10  & 0.07  & 0.03 \\ 
Cape Verde & 0.07  & 0.00  & 0.13  & 0.10  & 0.00  & 0.10  & 0.05  & 0.23  & 0.10  & 0.20  & 0.13  & 0.10  & 0.07  & 0.07  & 0.10  & 0.10  & 0.10  & 0.13  & 0.07  & 0.07  & 0.00 \\ 
Chadian & 0.00  & 0.03  & 0.07  & 0.15  & 0.13  & 0.13  & 0.07  & 0.30  & 0.20  & 0.20  & 0.13  & 0.30  & 0.10  & 0.13  & 0.20  & 0.23  & 0.13  & 0.13  & 0.10  & 0.17  & 0.10 \\ 
Comoros & 0.00  & 0.00  & 0.13  & 0.00  & 0.00  & 0.03  & 0.03  & 0.00  & 0.03  & 0.00  & 0.03  & 0.03  & 0.03  & 0.10  & 0.03  & 0.07  & 0.10  & 0.10  & 0.00  & 0.00  & 0.10 \\ 
Congo (Brazzaville) & 0.13  & 0.05  & 0.10  & 0.10  & 0.15  & 0.10  & 0.17  & 0.10  & 0.27  & 0.10  & 0.20  & 0.10  & 0.20  & 0.17  & 0.25  & 0.20  & 0.23  & 0.17  & 0.10  & 0.10  & 0.15 \\ 
Congo (Republic) & 0.00  & 0.07  & 0.10  & 0.00  & 0.17  & 0.20  & 0.03  & 0.07  & 0.07  & 0.10  & 0.03  & 0.20  & 0.17  & 0.03  & 0.20  & 0.23  & 0.13  & 0.03  & 0.10  & 0.10  & 0.25 \\ 
Côte d'Ivoire & 0.00  & 0.03  & 0.07  & 0.07  & 0.03  & 0.17  & 0.07  & 0.30  & 0.03  & 0.17  & 0.03  & 0.07  & 0.10  & 0.07  & 0.13  & 0.03  & 0.10  & 0.07  & 0.00  & 0.07  & 0.07 \\ 
Djibouti & 0.05  & 0.07  & 0.10  & 0.10  & 0.13  & 0.13  & 0.17  & 0.30  & 0.13  & 0.23  & 0.13  & 0.23  & 0.23  & 0.20  & 0.20  & 0.25  & 0.10  & 0.17  & 0.20  & 0.13  & 0.13 \\ 
Egypt & 0.10  & 0.00  & 0.03  & 0.10  & 0.03  & 0.07  & 0.00  & 0.20  & 0.03  & 0.17  & 0.03  & 0.10  & 0.13  & 0.07  & 0.03  & 0.07  & 0.00  & 0.07  & 0.15  & 0.10  & 0.07 \\ 
Equatorial Guinea & 0.00  & 0.00  & 0.03  & 0.00  & 0.07  & 0.10  & 0.00  & 0.07  & 0.00  & 0.10  & 0.03  & 0.05  & 0.00  & 0.00  & 0.05  & 0.05  & 0.00  & 0.00  & 0.00  & 0.07  & 0.05 \\ 
Eritrea & 0.10  & 0.00  & 0.03  & 0.07  & 0.03  & 0.07  & 0.07  & 0.20  & 0.03  & 0.03  & 0.07  & 0.13  & 0.03  & 0.03  & 0.13  & 0.17  & 0.07  & 0.13  & 0.03  & 0.03  & 0.07 \\ 
Ethiopia & 0.13  & 0.07  & 0.07  & 0.03  & 0.03  & 0.03  & 0.00  & 0.27  & 0.10  & 0.10  & 0.07  & 0.07  & 0.03  & 0.10  & 0.05  & 0.10  & 0.03  & 0.07  & 0.07  & 0.10  & 0.00 \\ 
Gabonese & 0.00  & 0.05  & 0.00  & 0.00  & 0.07  & 0.15  & 0.07  & 0.47  & 0.07  & 0.10  & 0.00  & 0.10  & 0.15  & 0.00  & 0.15  & 0.00  & 0.00  & 0.13  & 0.03  & 0.10  & 0.00 \\ 
Gambia & 0.13  & 0.03  & 0.10  & 0.03  & 0.10  & 0.17  & 0.10  & 0.17  & 0.13  & 0.35  & 0.17  & 0.23  & 0.17  & 0.27  & 0.13  & 0.17  & 0.23  & 0.17  & 0.13  & 0.17  & 0.10 \\ 
Ghana & 0.00  & 0.03  & 0.07  & 0.03  & 0.07  & 0.13  & 0.03  & 0.27  & 0.05  & 0.27  & 0.13  & 0.17  & 0.10  & 0.03  & 0.03  & 0.10  & 0.00  & 0.10  & 0.03  & 0.03  & 0.05 \\ 
Guinea & 0.07  & 0.03  & 0.07  & 0.10  & 0.10  & 0.20  & 0.10  & 0.53  & 0.13  & 0.20  & 0.17  & 0.10  & 0.20  & 0.23  & 0.10  & 0.23  & 0.05  & 0.17  & 0.07  & 0.20  & 0.13 \\ 
Guinea-Bissau & 0.10  & 0.00  & 0.10  & 0.03  & 0.03  & 0.07  & 0.03  & 0.17  & 0.10  & 0.07  & 0.05  & 0.30  & 0.10  & 0.20  & 0.13  & 0.25  & 0.20  & 0.13  & 0.03  & 0.03  & 0.03 \\ 
Kenya & 0.03  & 0.05  & 0.10  & 0.00  & 0.15  & 0.20  & 0.10  & 0.37  & 0.17  & 0.20  & 0.13  & 0.17  & 0.13  & 0.05  & 0.10  & 0.05  & 0.10  & 0.10  & 0.03  & 0.07  & 0.07 \\ 
Lesotho & 0.17  & 0.00  & 0.00  & 0.00  & 0.00  & 0.00  & 0.00  & 0.33  & 0.00  & 0.17  & 0.25  & 0.00  & 0.00  & 0.00  & 0.17  & 0.00  & 0.17  & 0.00  & 0.17  & 0.17  & 0.25 \\ 
Liberia & 0.00  & 0.00  & 0.00  & 0.04  & 0.19  & 0.04  & 0.16  & 0.19  & 0.08  & 0.08  & 0.12  & 0.08  & 0.04  & 0.12  & 0.12  & 0.00  & 0.12  & 0.04  & 0.00  & 0.00  & 0.08 \\ 
Libya & 0.00  & 0.12  & 0.19  & 0.28  & 0.21  & 0.18  & 0.13  & 0.33  & 0.10  & 0.46  & 0.24  & 0.32  & 0.13  & 0.10  & 0.16  & 0.19  & 0.22  & 0.21  & 0.12  & 0.12  & 0.13 \\ 
Madagascar & 0.07  & 0.04  & 0.10  & 0.04  & 0.00  & 0.07  & 0.10  & 0.28  & 0.10  & 0.24  & 0.15  & 0.11  & 0.17  & 0.07  & 0.17  & 0.10  & 0.07  & 0.18  & 0.00  & 0.07  & 0.04 \\ 
Malawi & 0.00  & 0.07  & 0.03  & 0.03  & 0.00  & 0.03  & 0.13  & 0.10  & 0.10  & 0.10  & 0.13  & 0.03  & 0.10  & 0.13  & 0.17  & 0.03  & 0.17  & 0.00  & 0.07  & 0.05  & 0.05 \\ 
Mali & 0.07  & 0.07  & 0.00  & 0.03  & 0.07  & 0.20  & 0.03  & 0.17  & 0.07  & 0.40  & 0.17  & 0.20  & 0.17  & 0.17  & 0.10  & 0.13  & 0.10  & 0.10  & 0.07  & 0.03  & 0.07 \\ 
\bottomrule
\end{tabular}%
}
\caption{Detailed breakdown of average performance scores for each model across specific countries. Best viewed digitally due to the large number of entries.}
\label{tab:t1}
\end{table*}

\end{document}